\documentclass[10pt,journal,compsoc]{IEEEtran}

\ifCLASSOPTIONcompsoc
  \usepackage[nocompress]{cite}
\else
  \usepackage{cite}
\fi
\usepackage{graphicx}
\usepackage{amsmath}
\usepackage{amssymb}
\usepackage{booktabs}
\usepackage{bbding}
\usepackage{epsfig}
\usepackage{multirow}
\usepackage{threeparttable}  
\usepackage{color}
\usepackage{array}
\usepackage{makecell}
\usepackage{stfloats}
\usepackage[table]{xcolor}
\usepackage[misc]{ifsym}
\usepackage[accsupp]{axessibility}
\usepackage{hyperref}
\usepackage{subfigure}

\newcommand{\ie}{\textit{i.e.}}

\ifCLASSINFOpdf
\else
\fi

\begin{document}

\title{JointFormer: A Unified Framework with Joint Modeling for Video Object Segmentation}

\author{Jiaming~Zhang, Yutao~Cui,
Gangshan~Wu,~\IEEEmembership{Member,~IEEE} and~Limin~Wang,~\IEEEmembership{Senior Member,~IEEE}
\IEEEcompsocitemizethanks{\IEEEcompsocthanksitem Jiaming Zhang, Yutao Cui, and Gangshan Wu are with the State Key Laboratory for Novel Software Technology, Nanjing University, Nanjing 210023, China (e-mail: jiamming.zhang@gmail.com; cuiyutao@smail.nju.edu.cn; gswu@nju.edu.cn).
\IEEEcompsocthanksitem Limin Wang is with the State Key Laboratory for Novel Software Technology, Nanjing University, Nanjing 210023, China, and also with the Shanghai Artificial Intelligence Laboratory, Shanghai 200232, China (e-mail: lmwang@nju.edu.cn).
}
}


\IEEEtitleabstractindextext{%
\begin{abstract}
Current prevailing Video Object Segmentation methods follow the pipeline of extraction-then-matching, which first extracts features on current and reference frames independently, and then performs dense matching between them.
This decoupled pipeline limits information propagation between frames to high-level features, hindering fine-grained details for matching.
Furthermore, the pixel-wise matching lacks holistic target understanding, making it prone to disturbance by similar distractors.
To address these issues, we propose a unified VOS framework, coined as~\emph{\bf JointFormer}, for jointly modeling feature extraction, correspondence matching, and a compressed memory.
The core Joint Modeling Block leverages attention to simultaneously extract and propagate the target information from the reference frame to the current frame and a compressed memory token.
This joint scheme enables extensive multi-layer propagation beyond high-level feature space and facilitates robust instance-distinctive feature learning.
To incorporate the long-term and holistic target information, we introduce a compressed memory token with a customized online updating mechanism, which aggregates target features and facilitates temporal information propagation in a frame-wise manner, enhancing global modeling consistency.
Our JointFormer achieves a new state-of-the-art performance on the DAVIS 2017 val/test-dev (89.7\% and 87.6\%) benchmarks and the YouTube-VOS 2018/2019 val (87.0\% and 87.0\%) benchmarks, outperforming the existing works.
To demonstrate the generalizability of our model, it is further evaluated on four new benchmarks with various difficulties, including MOSE for complex scenes, VISOR for egocentric videos, VOST for complex transformations, and LVOS for long-term videos.
Without specific design or modification to address these unusual difficulties, our model achieves the best performance across all benchmarks when compared with several current best models, illustrating its excellent generalization and robustness.
Further extensive ablations and visualizations indicate our compact JointFormer enables more comprehensive and effective feature learning and matching.
Code and trained models are available at \url{https://github.com/MCG-NJU/JointFormer}.

\end{abstract}

}

\maketitle

\IEEEdisplaynontitleabstractindextext

\IEEEpeerreviewmaketitle

\ifCLASSOPTIONcompsoc
\IEEEraisesectionheading{\section{Introduction}\label{sec:introduction}}
\else
\section{Introduction}
\label{sec:introduction}
\fi

\IEEEPARstart{S}{emi}-supervised Video Object Segmentation (VOS) is a challenging task in computer vision with many potential applications~\cite{ngan2011video,Zhang_2016_CVPR}, such as interactive video editing, video inpainting, and autonomous driving. It aims to track and segment the object(s) throughout a video sequence based on the mask annotation given in the first frame only.
Due to the limited target information provided, there comes out a core problem of \emph{how to capture discriminative \textcolor{black}{target-relevant} representation and propagate the target information at different levels for accurate \textcolor{black}{video} object segmentation}.

\begin{figure}[t]
\begin{center}
\includegraphics[width=\linewidth]{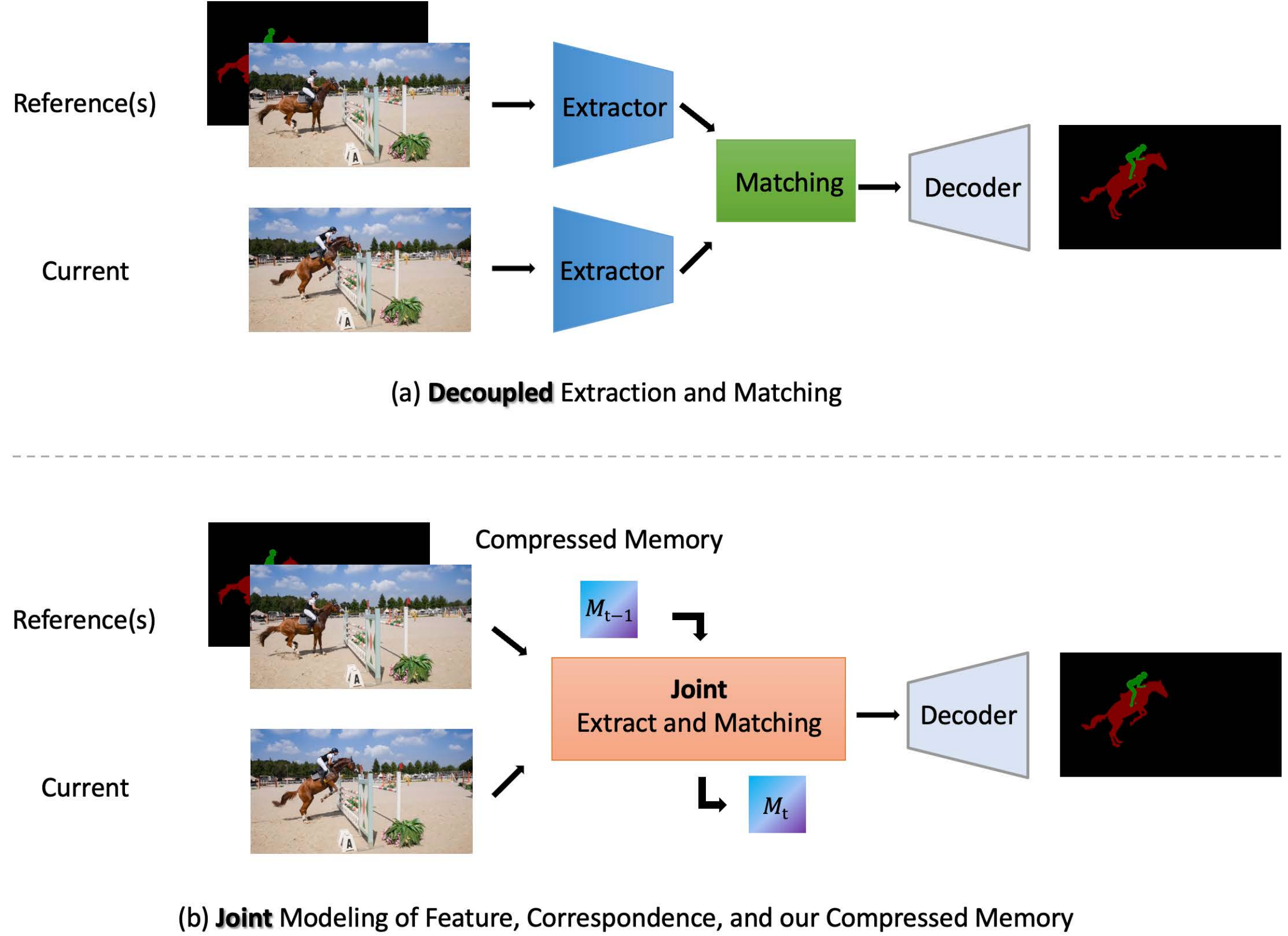}
\end{center}
\caption{
The pipeline comparison between existing VOS works (a) and ours (b).
(a) Existing works perform feature extraction and matching separately in a decoupled way.
(b) Our framework jointly models feature, correspondence, and our compressed memory without the post-matching in a unified pipeline.
}
\vspace{-4mm}
\label{fig:pipeline_comparison}
\end{figure}

Unlike classical image segmentation tasks, \textcolor{black}{which only operate within frames}, the key of the VOS task is how to propagate the target information, \ie, learning the target appearance and characteristics from \textcolor{black}{the first frame with its mask} and passing it frame by frame to \textcolor{black}{identify target} and adapt to its appearance variations.
The propagation-based methods~\cite{MaskTrack_Perazzi_2017_CVPR,RGMP_Oh_2018_CVPR,RMNet_Xie_2021_CVPR,Yang_2018_CVPR,SegFlow_Cheng_2017_ICCV,Li_2018_ECCV,Tsai_2016_CVPR} give their answers by iteratively propagating the masks with temporal correlations.
Besides, the prevailing matching-based methods~\cite{ PWL_Chen_2018_CVPR,VideoMatch_Hu_2018_ECCV,CFBI_10.1007/978-3-030-58558-7_20,FEELVOS_Voigtlaender_2019_CVPR} pursue a various \textcolor{black}{kinds} of performing dense matching between the top-level feature of current and reference frames by calculating the correspondence map.
Furthermore, several memory-based methods~\cite{STM_Oh_2019_ICCV,KMN_10.1007/978-3-030-58542-6_38,AFBURR_NEURIPS2020_23483314,STCN_NEURIPS2021_61b4a64b,XMem_10.1007/978-3-031-19815-1_37} leverage a memory bank to store the multiple reference frames with \textcolor{black}{their} masks as spatially fine-grained memory \textcolor{black}{to diversify the matches to the current features}.
Despite the significant success they have achieved, there still exist the following drawbacks.
\romannumeral1) They all follow a fixed \textbf{extract-then-matching} pipeline as shown in Fig~\ref{fig:pipeline_comparison}(a), \ie, first \emph{extracting} the feature of current and reference frames with backbone independently, and then performing \emph{matching} only at high-level feature space to propagate the target information.
On one hand, the decoupled feature extractor and matching module inevitably makes the model struggle to capture target-specific features at the lower levels, and the features used for matching are fixed and target-independent, which \textcolor{black}{are harmful for} fine-grained segmentation and discriminative learning.
On the other hand, the \textcolor{black}{specially designed and untrained} integration module lacks the flexibility of enjoying the development of large-scale pre-training, such as the Masked Image Modeling (MIM)~\cite{MAE_He_2022_CVPR,ConvMAE_gao2022convmae}.
\romannumeral2) Typically, the matching is processed in a pixel-wise way, \ie, performing dense propagation between all elements in reference features and the current features \textcolor{black}{independently}.
The modeling of holistic targets, however, tends to be overlooked, which may lead to deficient discriminative capability and incorrect matching with distractors.

To address the above issues, we bring a new perspective to the VOS task that the modeling of feature, fine-grained correspondence and instance-level compressed memory should be coupled in a compact transformer architecture (refer to Fig.~\ref{fig:pipeline_comparison}).
Such joint modeling \textcolor{black}{scheme}, a concise and unified design, can bring us several key benefits.
First, the joint modeling can unlock the potential of capturing extensive and discriminative target-specific features, unleash the strong power of MIM pre-training for all three processes and \textcolor{black}{assist} each other.
Second, unlike existing matching which provides spatially fine-grained features at the \textcolor{black}{pixel-level}, our compressed memory represents each object as a whole instance, so as to provide a comprehensive and discriminative understanding for the objects and adapt to the deformation of objects with \textcolor{black}{our temporal updating design}.

Driven by the above analysis, we propose a unified framework that jointly models feature extraction, correspondence matching, and our compressed memory, coined as \textbf{JointFormer}.
Specifically, we first \textcolor{black}{tokenize} the current frame and reference frames with masks, concatenate them with the compressed memory embedding into token sequences and feed them into stacks of transformer-based \textbf{Joint Modeling Block}.
With the help of the attention mechanism, the Joint Modeling Blocks perform iterative modeling to propagate the target information in multiple flows, hence achieving the above goals.
Due to the special qualities of reference and current in the VOS task, we also design multiple modes of information propagation and in-depth investigate \textcolor{black}{their various impacts}.
Especially, the presented \textbf{compressed memory} stores only one token for each target, so as to represent it at the instance level.
To incorporate the long-term temporal target information, we design a customized online updating mechanism for the compressed memory token. 
In detail, we take the multi-level tokens, generated from the backbone or decoder in the previous frame, as the temporal memory for the compressed token \textcolor{black}{to update it}, which enables the target information to propagate along the temporal dimension.
Under these designs, we formulate a compact, unified and concise VOS framework, allowing for accurate and robust video object segmentation.

Our contributions can be summarized as follows:
\begin{itemize}
\item We propose a unified network, coined as \emph{JointFormer}, to jointly model feature extraction, correspondence matching, and a compressed memory. The compact and \textcolor{black}{unified} framework enables extensive information propagation and discriminative learning.
\item We develop a customized online updating mechanism for the compressed memory, which helps to \textcolor{black}{aggregate target feature as a whole instance and endow with
temporal information propagation in frame-wise manner to improve the global modeling consistency}.
\item Comprehensive experiments show that our \emph{JointFormer} achieves state-of-the-art performance with 89.7\% and 87.6\% on DAVIS 2017~\cite{DAVIS17_Pont-Tuset_arXiv_2017} validation and test-dev split, 87.0\% and 87.0\% on YouTube-VOS~\cite{youtubeVOS_xu2018youtube} 2018 \& 2019 validation split. In addition, we test our JointFormer on four new benchmarks with various difficulties, including MOSE, VOST, VISOR, and LVOS \textcolor{black}{with their respective settings. Compared} with several current best models, the results show that our method achieves the best performance on all four benchmarks, indicating that our model \textcolor{black}{can overcome various challenges and has a strong generalization}.
\end{itemize}

\section{Related Work}

\subsection{Semi-supervised VOS}
The Semi-supervised VOS task aims to track and segment object(s) in the video sequence based on its mask given in the first frame.
Early online learning-based methods~\cite{OSVOS_Caelles_2017_CVPR,MoNet_Xiao_2018_CVPR,OnAVOS_BMVC2017_116,DBLP:journals/corr/KhorevaBIBS17} rely on fine-tuning pre-trained segmentation networks at test time to make the networks focus on the given object.
OSVOS~\cite{OSVOS_Caelles_2017_CVPR} uses a pre-trained object segmentation network, then fine-tunes networks on the first-frame annotation at test time, OnAVOS~\cite{OnAVOS_BMVC2017_116} extends it with an online adaptation mechanism.
MaskTrack~\cite{MaskTrack_Perazzi_2017_CVPR} and PReM~\cite{PReM_10.1007/978-3-030-20870-7_35} utilize optical flow to propagate the segmentation mask from one frame to the next.
Despite achieving promising results, they are slowed down due to the limitations of fine-tuning on the first frame during inference.
Propagation-based methods~\cite{ MaskTrack_Perazzi_2017_CVPR,RGMP_Oh_2018_CVPR,RMNet_Xie_2021_CVPR,Yang_2018_CVPR,SegFlow_Cheng_2017_ICCV,Li_2018_ECCV,Tsai_2016_CVPR} formulate this problem as an object mask propagation task and iteratively propagate the segmentation masks with temporal correlations, but they are prone to drifting and struggle with occlusions.
Matching-based methods~\cite{PWL_Chen_2018_CVPR,VideoMatch_Hu_2018_ECCV,CFBI_10.1007/978-3-030-58558-7_20,FEELVOS_Voigtlaender_2019_CVPR,CFBIP_yang2020} calculate the correspondence pixel map between the current frame and the reference frame \textcolor{black}{with various correlation operations}.
PWL\cite{PWL_Chen_2018_CVPR} learns pixel-wise embedding with a nearest neighbor classifier via pixel-wise metric learning.
VideoMatch\cite{VideoMatch_Hu_2018_ECCV} uses a soft matching layer to compute a pixel-wise correlation map in a learned embedding space.
FEELVOS\cite{FEELVOS_Voigtlaender_2019_CVPR} and CFBI\cite{CFBI_10.1007/978-3-030-58558-7_20, CFBIP_yang2020} add \textcolor{black}{extra} local matching with the previous frame beyond matching with the first frame to adapt to current changes in targets.
Recent memory-based methods~\cite{STM_Oh_2019_ICCV,KMN_10.1007/978-3-030-58542-6_38,AFBURR_NEURIPS2020_23483314,STCN_NEURIPS2021_61b4a64b,JOINT_Mao_2021_ICCV,MiVOS_Cheng_2021_CVPR,XMem_10.1007/978-3-031-19815-1_37} propose an external memory bank to store the past frames in order to address the context limitation and design various memory storage and rating criteria.
Some works~\cite{SSTVOS_Duke_2021_CVPR, TransVOS_mei2021transvos, AOT_NEURIPS2021_147702db,DeAOT_yang2022decoupling} adopt transformer blocks~\cite{Transformer_NIPS2017_3f5ee243} for better matching, but train the blocks from scratch without pre-training.
SST\cite{SSTVOS_Duke_2021_CVPR} uses the transformer’s encoder with sparse attention to calculate pixel-level matching maps,
TransVOS\cite{TransVOS_mei2021transvos} introduces a vision transformer to model both the temporal and spatial relationships,
AOT~\cite{AOT_NEURIPS2021_147702db} and DeAOT~\cite{DeAOT_yang2022decoupling} propose a Long Short-Term Transformer structure for constructing hierarchical propagation for tracking and segmenting multiple objects uniformly and simultaneously.
Despite the promising results, they all follow a fixed pipeline that extracts features of the current and reference frames separately and then \textcolor{black}{match} them, limiting the \textcolor{black}{target-relevant} information of current feature.
Unlike them, we jointly model features and correspondence inside the full vision transformer~\cite{vit,ConvMAE_gao2022convmae}, allowing them to help each other, and minimize its modification to \textcolor{black}{maximize unleash the power of backbone pre-training}.
\textcolor{black}{TCOW\cite{tcow} adopts video backbone Timesformer\cite{timesformer} to interact frames together with masks a tube. Although the joint modeling approach is much like ours, it employs a simple and fully self-attention mechanism without specifying the reference frame and the current frame. In contrast, our approach modifies the information interaction as an asymmetric structure to avoid interference from reference frames, and compressed memory to propagate holistic target features across frames.
\textcolor{black}{
SimVOS~\cite{SimVOS}, which is our concurrent method, also performs joint feature extraction and matching inside a transformer-based backbone as we do.
However, it adopts a simple self-attention mechanism by fully interacting the reference and current features, causing the reference features to be affected by non-target regions in the current features.
In contrast, beyond the unified transformer-based framework, we consider the specificity of the VOS task with unique design, including designing the interaction as an asymmetric structure that prevents the target information of the reference features from being affected by current features, and compressed memory that is jointly modeled inside the unified framework and compresses the target features into a single token in order to treat the target as a whole instance.
The experimental results show that our method outperforms SimVOS in performance, demonstrating the effectiveness of our design.
}
VITA~\cite{VITA} adopts frame query to hold object-centric information collected throughout the whole video, and video query to aggregate information from the frame queries. Both of queries are initialized for each frame to operate independently without any cross-frame propagation. Our compressed memory aggregates the target features for each frame and then passes them to the next frame for feedback, enabling cross-frame target feature propagation.
}

\subsection{Memory Bank and Target Feature}
How to design the memory bank and what needs to be stored is critical for memory-based methods.
But most works~\cite{STM_Oh_2019_ICCV,STCN_NEURIPS2021_61b4a64b} choose to only memorize the reference frames with masks to provide fine-grained information in pixel-level.
AFB-URR~\cite{AFBURR_NEURIPS2020_23483314} and QDMN~\cite{QDMN_10.1007/978-3-031-19818-2_27} dynamically manage them according to the similarity or segmentation quality.
However, these works lack a comprehensive understanding of the target, making it difficult to solve the large deformation problem and distinguish between similar objects at the pixel level.
XMem~\cite{XMem_10.1007/978-3-031-19815-1_37} introduces multi-store memory and dynamic \textcolor{black}{updates} its sensory memory with Gated Recurrent Unit (GRU)~\cite{GRU1_cho2014learning}, but it can only be updated using the top-level feature and the additional temporal module needs to be trained from scratch.
Another several methods choose to \textcolor{black}{summarizes} the target feature directly.
HODOR~\cite{HODOR} and ISVOS~\cite{ISVOS_Wang2022LookBY} encode object feature into descriptors (or object queries) with a Transformer Decoder.
Again, they only interact with the top-level feature in one frame and the extra modules bring more parameters without pre-trained, and the latter \textcolor{black}{method} joins another dataset as supervised signals for the module.
AOT and DeAOT~\cite{AOT_NEURIPS2021_147702db,DeAOT_yang2022decoupling} project the mask of multiple objects together into identification embeddings with an ID matrix and add them to the value of attention.
However, they only compress the mask \textcolor{black}{with its original size into its ID embeddings}, which neither \textcolor{black}{leverages the image representation} nor utilizes the multi-level features, \textcolor{black}{so that it} loses the object information severely.
Unlike them, our compressed memory views each target as a whole instance during the entire video. The unified framework jointly models pixel-level mask memory and our instance-level compressed memory inside the backbone and makes them conceptually justified and compensate for each other.
On the one hand, since \textcolor{black}{the customized online updating mechanism for our compressed memory is performed} inside the backbone, the compressed memory can learn multi-level features to overcome the object deformation problem.
On the other hand, \textcolor{black}{it} provides long-term features passed between frames through  \textcolor{black}{the customized online updating mechanism}, which is suitable for online video tasks.

\subsection{Joint Learning and Relation in SOT Task}\label{sec:relation_with_sot}
Recently, thanks to the parallelization and performance, Transformer~\cite{Transformer_NIPS2017_3f5ee243} was introduced to many computer vision tasks, such as image classification~\cite{vit, CvTWu_2021_ICCV, PvT_Wang_2021_ICCV}, object detection ~\cite{DETR_10.1007/978-3-030-58452-8_13, DeformableDETR_zhu2021deformable, DN-DETR_Li_2022_CVPR}, semantic segmentation~\cite{SETR_Zheng_2021_CVPR, SegFormer_NEURIPS2021_64f1f27b, SegViT_zhang2022segvit}.
In the single object tracking (SOT) task, the dominant tracking framework follows the \textcolor{black}{three-stage} pipeline that first extracts the features of the \textcolor{black}{templates} and search region separately, performs relation modeling, \textcolor{black}{and locates the target with box head finally}.
In order to simplify this pipeline, some works~\cite{MixFormer_Cui_2022_CVPR,OSTrack_10.1007/978-3-031-20047-2_20,SimTrack_10.1007/978-3-031-20047-2_22} unify the process of feature extraction and target information integration simultaneously inside the Transformer backbone. With the help of the transformer structure and attention operation, their unified one-stream \textcolor{black}{structures} have achieved excellent performance.
Note that they almost work with the bidirectional information propagation, which means they concatenate the template frame and search frame as a whole sequence and perform self-attention to make them share information in a \textcolor{black}{bidirectional mode}.
DropMAE~\cite{DropMAE_Wu_2023_CVPR} redesign MAE\cite{MAE_He_2022_CVPR} to be pre-trained on videos for facilitating temporal correspondence learning for SOT task and build a simple VOS baseline following the bidirectional propagation with its pre-training.
\textcolor{black}{
The SOT and VOS tasks perform coarse- and fine-grained tracking of targets in the form of boxes and masks, respectively, so both of them emphasize target feature extraction and relationship modeling.
However, even though the task essence of both SOT and VOS is tracking, the two tasks are inconsistent in terms of the granularity of the tracking and the number of targets, which makes it difficult for one task to benefit directly from the other.
\romannumeral1) The template of SOT only provides an image of a single target with its bounding box so that both sides of its interaction are equivalent pictures.
In contrast to SOT, the reference of VOS contains extra non-RGB input, \emph{i.e}, the masks of multiple targets, making the interaction non-equivalent.
Furthermore, the non-target region of the reference frame is filtered through the mask, while the current frame is retained entirely, which interferes with the target features of the reference frame with the bidirectional interaction of SOT.
\romannumeral2) Unlike the SOT task, which tracks only one target, most of the videos in the VOS task are multi-target.
It is more important to provide a holistic and discriminative feature for each target to distinguish similar objects.
Combining the above analysis, we first modify the bidirectional structure of interaction to protect reference from the distractions of non-target areas of current frames.
}
To further provide discriminability of various similar targets, we present a compressed memory \textcolor{black}{with a customized online updating mechanism} that provides a long-term and adaptable feature \textcolor{black}{for each target by regarding it as a whole instance beyond pixel-independent matching}.

\section{Method}

\begin{figure*}[pt]
\begin{center}
\includegraphics[width=\linewidth]{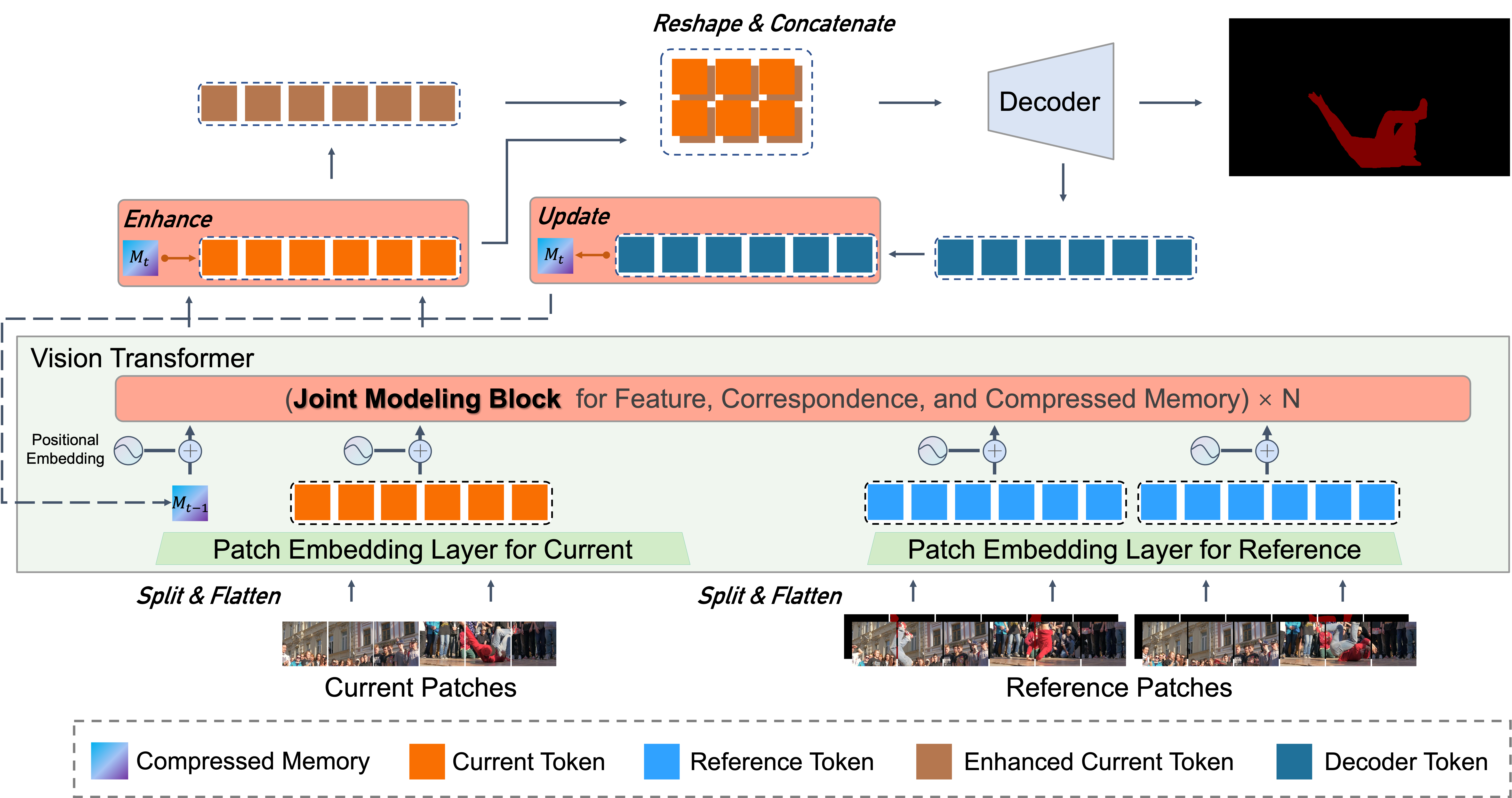}
\end{center}
\vspace{-3mm}
\caption{
Overview of our \textbf{JointFormer}.
The current frames and reference frames with masks are split and flattened into patches,
\textcolor{black}{then are fed} with our \textbf{compressed memory} into the Vision Transformer, which consists of \textbf{Joint Modeling Blocks} \textcolor{black}{for simultaneous feature extraction and target information propagation from the reference features to the current features and the compressed memory token.
After that, the current tokens are enhanced with compressed memory, reshaped to 2D shape, concatenated with enhanced current tokens, and then fed into the decoder for mask prediction.
In addition, the decoder fuses its internal features with mask prediction to generate decoder tokens for further updating the compressed memory token, and is subsequently used for predicting the next frame.
}
}
\label{fig:arch_overview}
\vspace{-3mm}
\end{figure*}

\subsection{Revisit the Previous Decoupled Extraction and Matching \textcolor{black}{Pipeline}}

We begin with a review of the previous prevailing \textcolor{black}{pipeline}, as shown in Fig.~\ref{fig:pipeline_comparison}(a), and consider a single object for convenience.
\textcolor{black}{
Suppose there are $T$ reference frames, each with corresponding masks, and one current frame.
The model first input reference frames $I_{z} \in \mathbb{R}^{3 \times THW}$ and current frame $I_{x} \in \mathbb{R}^{3 \times HW}$ into Key Encoder respectively to \textbf{extract} their keys $K_{z} \in \mathbb{R}^{C^{k} \times THW}$ and $K_{x} \in \mathbb{R}^{C^{k} \times HW}$. Meanwhile, the frames and masks of references are concatenated as $R_{z} \in \mathbb{R}^{4 \times THW}$ then fed into Value Encoder to \textbf{extract} value $V_{z} \in \mathbb{R}^{C^{v} \times THW}$.
After that, they perform \textbf{feature matching} that calculates affinity matrix $W_{z, x} \in \mathbb{R}^{THW \times HW}$ between the two keys with various similarity functions, then reads out the target-specific feature $F_{x} \in \mathbb{R}^{C^{v} \times HW}$ from the value of references through weighted sum function $F_{x} = V_{R}W$.
}
Finally, the target-specific feature $F_{x}$ is passed into the decoder for mask prediction $M_{x} \in \mathbb{R}^{1 \times HW}$.
\textcolor{black}{Although the previous method was intuitive}, the \textbf{extract-then-matching} pipeline restricts the capture of target-specific features at the lower levels, and the pixel-level matching neglects the holistic understanding of the target at the instance level.

\subsection{Architecture Overview}

In this section, we present a neat, unified network \textbf{JointFormer} as illustrated in Fig.~\ref{fig:arch_overview}.
Firstly, we concatenate $T$ reference frames with their masks along channel dimension to obtain reference pairs $R_{z} \in \mathbb{R}^{4 \times THW}$.
\textcolor{black}{
Then we split and flatten the reference pairs $R_{z}$ and current frame $I_{x}$ into patches $\boldsymbol{p}_{z} \in \mathbb{R}^{N_{z} \times (4 \cdot P^{2})}$ and $\boldsymbol{p}_{x} \in \mathbb{R}^{N_{x} \times (3 \cdot P^{2})}$, where $P \times P$ is the resolution of each patch, and $N_{z}=THW/P^{2}$, $N_{x}=HW/P^{2}$ are the number of patches of references and current frames respectively. After that, two trainable linear projection layers with parameter $\boldsymbol{E}_{z} \in \mathbb{R}^{(4 \cdot P^{2}) \times D}$ and $\boldsymbol{E}_{x} \in \mathbb{R}^{(3 \cdot P^{2}) \times D}$ are used to project $\boldsymbol{p}_{z}$ and $\boldsymbol{p}_{x}$ into patch embeddings in $D$ dimension space. To interpolate the positional correlation between patch embeddings, the learnable 1D position embeddings $\boldsymbol{P}_{z} \in \mathbb{R}^{N_{z} \times 1}$ and $\boldsymbol{P}_{x} \in \mathbb{R}^{N_{x} \times 1}$ are added to the patch embeddings of references and current separately to produce the reference token embeddings $\boldsymbol{H}_{z}^{0} \in \mathbb{R}^{N_{z} \times D}$ and current token embeddings $\boldsymbol{H}_{x}^{0} \in \mathbb{R}^{N_{x} \times D}$, which can be written as:
}

\color{black}
\begin{equation}
\begin{aligned}
    \boldsymbol{H}_{z}^{0} &= [\boldsymbol{p}_{z}^{1}\boldsymbol{E}_{z}; \boldsymbol{p}_{z}^{2}\boldsymbol{E}_{z}; \dots ; \boldsymbol{p}_{z}^{N_{z}}\boldsymbol{E}_{z}] + \boldsymbol{P}_{z}, \\
    \boldsymbol{H}_{x}^{0} &= [\boldsymbol{p}_{x}^{1}\boldsymbol{E}_{x}; \boldsymbol{p}_{x}^{2}\boldsymbol{E}_{x}; \dots ; \boldsymbol{p}_{x}^{N_{x}}\boldsymbol{E}_{x}] + \boldsymbol{P}_{z}. \\
\end{aligned}
\label{eq:pe}
\end{equation}
\color{black}

\textcolor{black}{
The two token embeddings $\boldsymbol{H}_{x}^{0}$ and $\boldsymbol{H}_{z}^{0}$ are then concatenated with our compressed memory token $\boldsymbol{H}_{M}^{0} \in \mathbb{R}^{1 \times D}$ as $\boldsymbol{H}_{Mzx}^{0}=[\boldsymbol{H}_{M}^{0}; \boldsymbol{H}_{z}^{0}; \boldsymbol{H}_{x}^{0}]$, and then the concatenated token sequence $\boldsymbol{H}_{Mzx}^{0}$ is fed into Transformer-based backbone, which consists of our \textbf{Joint Modeling Blocks}} to propagate the target information, achieving jointly model feature, correspondence, and our \textbf{compressed memory} simultaneously inside the backbone, which is different from the previous pipeline.
\textcolor{black}{After the operation of $L$ Joint Modeling Blocks, the token sequence is decoupled as our compressed memory token $\boldsymbol{H}_{M}^{L}$, reference tokens $\boldsymbol{H}_{z}^{L}$, and current token $\boldsymbol{H}_{x}^{L}$, where the current tokens are target-specific features and compressed memory token has aggregated target information as a whole instance. The current tokens are then \textbf{enhanced} by the holistic target feature contained in compressed memory as $\bar{\boldsymbol{H}}_{x}^{L}$. The two current token sequences are reshaped to 2D shape and concatenated along their channel dimension, then fed it with multi-scale feature into} the decoder to predict the mask $\boldsymbol{M}_{x} \in \mathbb{R}^{1 \times HW}$. \textcolor{black}{At the same time, the decoder fuses its internal feature and logits prediction to propose decoder tokens $\boldsymbol{H}_{D} \in \mathbb{R}^{N_{x} \times D}$ for \textbf{updating} the compressed memory token with our proposed \textbf{customized online updating mechanism} to endow with temporal information propagation in a frame-wise manner to improve the global modeling consistency.}

\begin{figure}[t]
\begin{center}
\includegraphics[width=1.0\columnwidth,keepaspectratio]{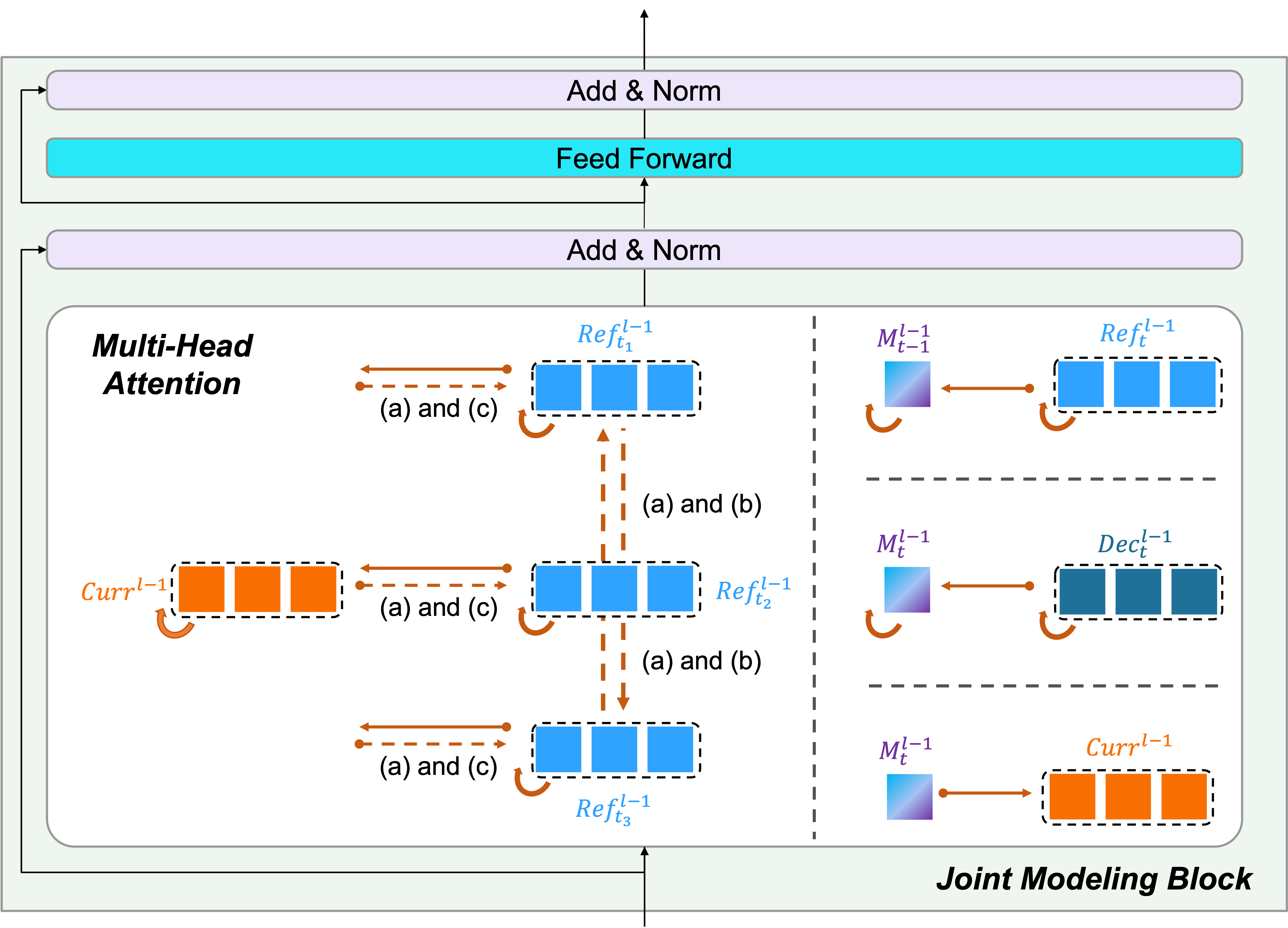}
\end{center}
\vspace{-3mm}
\caption{
\textcolor{black}{
Detailed view of the Joint Modeling Block, for joint modeling of features, correspondence, and compressed memory with attention mechanisms, shown as the \textcolor{brown}{brown arrow}.
Specifically, the source of the arrow represents the key/value and the pointer represents the query.
The dotted arrow indicates that interaction only exists in the side-labeled corresponding mode.}
}
\vspace{-5mm}
\label{fig:arch_joint_block}
\end{figure}

\subsection{Joint Modeling Block} \label{sec:method_joint_block}

The previous works perform feature extraction and relation matching separately, making it difficult to capture lower-level target representations and handle the large-scale pre-training.
Unlike them, we proposed \textbf{Joint Modeling Block}, as the core module of our JointFormer, joint modeling feature, correspondence, and our compressed memory.
As shown in Fig.~\ref{fig:arch_joint_block}, to \textcolor{black}{simultaneously extract features and propagate target information, the Joint Modeling Blocks adopt multi-head attention to model the information propagation. A straightforward way is to treat} the reference tokens and current tokens as a whole token sequence and perform self-attention on the sequence \textcolor{black}{so that the information propagation is bi-directional and fully shared.
Nevertheless, it should be noted that the reference tokens are generated from the reference frames and masks, and the non-target regions of references are well-filtered by masks.
In contrast, the current frame contains a large amount of non-target information, which would seriously disturb the target features of reference features in the case of bi-directional information propagation.
Based on the above analysis, the way of information propagation needs to be carefully designed to propagate the target information from the reference to the current feature while minimizing the disturbance from other features.
}
To further analyze the impact of information propagation, we propose four modes for propagating target information.
\textcolor{black}{Specifically, we modify the query and key-value pairs of multi-head attention, which is the core mechanism for propagation as following:}

\color{black}
\vspace{-4mm}
\begin{equation}
\begin{aligned}
    Q_{z_{ti}}, Q_{x} &= W^{Q}\boldsymbol{H}_{z_{ti}}^{l}, W^{Q}\boldsymbol{H}_{x}^{l}, \\
    K_{x}, V_{x} &= W^{K,V}[\boldsymbol{H}_{z_{t1}}^{l}; \boldsymbol{H}_{z_{t2}}^{l}; \boldsymbol{H}_{x^{l}}],\\
    \mathrm{Attention}_{x}^{l} &= \mathrm{softmax}(\frac{Q_{x}K_{x}^{T}}{\sqrt{d}})V_{x}, \\
    K_{z_{ti}}, V_{z_{ti}} &=
    \begin{cases}
    W^{K, V}[\boldsymbol{H}_{z_{ti}}^{l}, \boldsymbol{H}_{z_{t \neq ti}}^{l}, \boldsymbol{H}_{x^{l}}],  & \text{(a)} \\
    W^{K, V}[\boldsymbol{H}_{z_{ti}}^{l}, \boldsymbol{H}_{z_{t \neq ti}}^{l}], & \text{(b)} \\
    W^{K, V}[\boldsymbol{H}_{z_{ti}}^{l}, \boldsymbol{H}_{x^{l}}],  & \text{(c)} \\
    W^{K, V}[\boldsymbol{H}_{z_{ti}}^{l}], & \text{(d)} \\
    \end{cases} \\
    \mathrm{Attention}_{z_{ti}}^{l} &= \mathrm{softmax}(\frac{Q_{z_{ti}}K_{z_{ti}}^{T}}{\sqrt{d}})V_{z_{ti}}, \\
\end{aligned}
\label{eq:attn1}
\end{equation}
\color{black}
\textcolor{black}{where $W^{Q}$ and $W^{K,V}$ represent the projection of query and key-value in attention respectively, and $z_{ti}$ denotes the $i_{th}$ reference frame.}

\textcolor{black}{As shown in Figure.~\ref{fig:arch_joint_block}, in all four propagation modes, except that each feature is extracted from its own, the current feature interacts with all reference features.
In modes (a) and (b), the reference feature propagates information to other reference features while not in (c) and (d).
In modes (a) and (c), the current feature propagates to the reference feature while not in (b) and (d).
For more comprehensive analysis, we compare the performance of the four modes in Section~\ref{ablation:attention} and provide extensive visualizations to illustrate the impact of propagation.
}

\subsection{Compressed Memory} \label{sec:method_compressed_memory}

Most of the existing memory-based works rely only on reference masks, which are pixel-level memories. 
\textcolor{black}{However, the pixel-wise matching with these memories ignores the holistic modeling of the target, which results in failing to distinguish similar distractors and lacking a long-term understanding during the online segmentation process.}
To fix this issue and further provide discriminative target features, inspired by the "classification token" of ViT~\cite{vit} that can represent the entire image feature, we present \textbf{compressed memory} \textcolor{black}{$\boldsymbol{H}_{M}^{0} \in \mathbb{R}^{1 \times D}$}, just one token for each target to represent it as a whole instance.
Specifically, we design a \textbf{customized online updating mechanism} that \textcolor{black}{jointly models the compressed memory with the current tokens and reference tokens or decoder tokens in proposed Joint Modeling Blocks, as shown in Fig.~\ref{fig:arch_joint_block}.
After aggregating the target feature, the compressed memory \textbf{enhances} the target-specific information of current tokens.
}

\noindent \textbf{Customized online updating mechanism.}
\textcolor{black}{To provide a long-term and holistic feature, the compressed memory interacts with reference tokens at multiple layers as Eq.~\ref{eq:attn2}, enabling powerful instance-distinctive feature learning.
Furthermore, the compressed memory also interacts with the decoder token $\boldsymbol{H}_{D}$ together with its feature extraction, which is proposed by the decoder according to current feature and mask prediction, endowing with temporal information propagation in a frame-wise manner to improve the global modeling consistency.
In summary, the update mechanism is as follows:
}
\color{black}
\begin{equation}
\begin{aligned}
    Q_{z_{ti}}, Q_{M}, Q_{D} &= W^{Q}\boldsymbol{H}_{z_{ti}}^{L}, W^{Q}\boldsymbol{H}_{M}^{L}, W^{Q}\boldsymbol{H}_{D}, \\
    K_{M}, V_{M} &=
        \begin{cases}
        W^{K, V}[\boldsymbol{H}_{z_{ti}}^{L}, \boldsymbol{H}_{M}], \\
        W^{K, V}[\boldsymbol{H}_{D}, \boldsymbol{H}_{M}], \\
        \end{cases} \\
    \mathrm{Attention}_{M}^{L} &= \mathrm{softmax}(\frac{Q_{M}K_{M}^{T}}{\sqrt{d}})V_{M}, \\
    K_{D}, V_{D} &= W^{K,V}[\boldsymbol{H}_{D}], \\
    \mathrm{Attention}_{D} &= \mathrm{softmax}(\frac{Q_{D}K_{D}^{T}}{\sqrt{d}})V_{D}. \\
\end{aligned}
\label{eq:attn2}
\end{equation}
\color{black}

\noindent \textcolor{black}{\textbf{Target Enhancement.}}
After the updating mechanism, the compressed memory contains the comprehensive target feature.
We \textcolor{black}{leverage it to \textbf{enhance} target feature of current tokens with Joint Modeling Block} with a small modification.
\textcolor{black}{Specifically, due to only one key of the attention matrix between current tokens and compressed memory,} we modify $softmax$ in standard attention operation to $sigmoid$ to \textcolor{black}{increase the difference between the enhancement at various locations, which can be written as:}
\color{black}
\begin{equation}
\begin{aligned}
    Q_{x} &= W^{Q}\boldsymbol{H}_{x}^{L}, \\
    K_{x}, V_{x} &= W^{K,V}[\boldsymbol{H}_{M}^{L}],\\
    \bar{\boldsymbol{H}}_{x}^{L} &= \mathrm{sigmoid}(\frac{Q_{x}K_{M}^T}{\sqrt{d}})V_{M}. \\
\end{aligned}
\label{eq:attn3}
\end{equation}
\color{black}

Over a long period of frame-by-frame updating with Joint Modeling Blocks, our compressed memory \textcolor{black}{jointly modeled with reference features and decoder tokens at multiple layers,}
resulting in a stable and adaptable feature \textcolor{black}{to improve the global modeling consistency}.

\begin{figure}[t]
\begin{center}
\vspace{3mm}
\includegraphics[width=1.0\columnwidth,keepaspectratio]{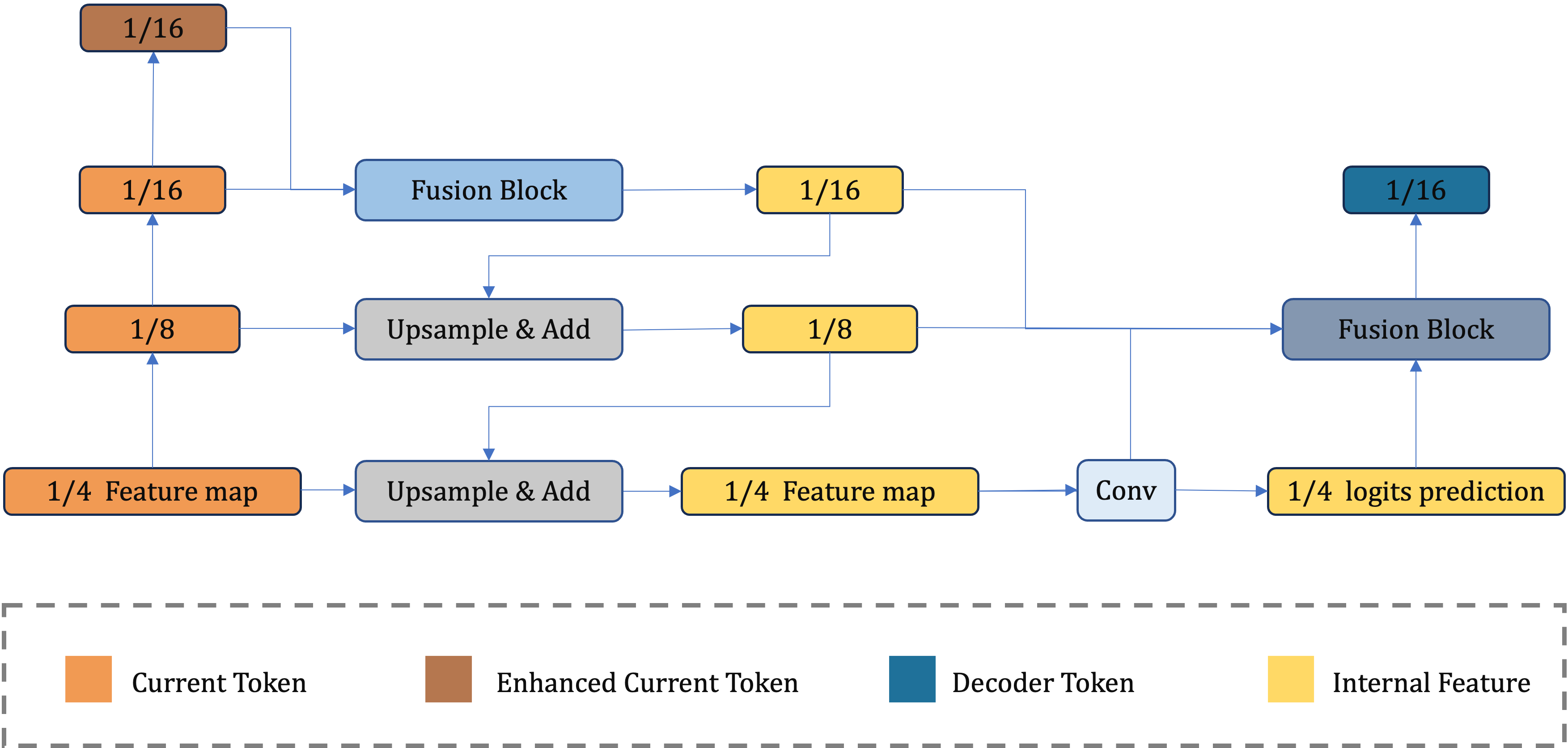}
\end{center}
\vspace{-3mm}
\caption{\textcolor{black}{Detailed view of the Decoder. We first fuse the current tokens and enhanced current tokens, then progressively upsample and fuse them with multi-scale features from backbone, and predict the target logits with a convolution finally.
Furthermore, we leverage the logits prediction and internal features within the decoder to generate decoder tokens for updating the compressed memory.}
}
\vspace{-5mm}
\label{fig:arch_decoder}
\end{figure}

\begin{table*}[t]
\caption{
Quantitative comparisons on the DAVIS 2017~\cite{DAVIS17_Pont-Tuset_arXiv_2017}, YouTube-VOS\cite{youtubeVOS_xu2018youtube} 2018 \& 2019 dataset.
The $^{\ast}$ and $^{\dag}$ denote pre-trained on image datasets and the large BL30K dataset\cite{MiVOS_Cheng_2021_CVPR}, respectively.
The results in \textcolor{blue}{blue} and \textcolor{red}{red} indicate the best of the previous and all works.
We re-run AOT~\cite{AOT_NEURIPS2021_147702db} and DeAOT~\cite{DeAOT_yang2022decoupling} on YouTube-VOS dataset \textcolor{black}{under} all input frames \textcolor{black}{setting} for fair comparison which improves their performance.
}
\begin{center}
\fontsize{9pt}{4mm}\selectfont
\setlength{\tabcolsep}{1.6mm}
\begin{tabular}{l|ccc|ccc|ccccc|ccccc}
\toprule[1.5pt]
\multirow{2}{*}{Method} & \multicolumn{3}{c|}{DAVIS 2017 val} & \multicolumn{3}{c|}{DAVIS 2017 test} & \multicolumn{5}{c|}{YouTube-VOS 2018 val} & \multicolumn{5}{c}{YouTube-VOS 2019 val} \\ \cline{2-17} 
\rule{0pt}{9pt} & $\mathcal{J}\&\mathcal{F}$       & $\mathcal{J}$          & $\mathcal{F}$         & $\mathcal{J}\&\mathcal{F}$       & $\mathcal{J}$          & $\mathcal{F}$          & $\mathcal{G}$     & $\mathcal{J}_S$    & $\mathcal{F}_S$    & $\mathcal{J}_U$    & $\mathcal{F}_U$   & $\mathcal{G}$     & $\mathcal{J}_S$    & $\mathcal{F}_S$    & $\mathcal{J}_U$   & $\mathcal{F}_U$   \\
\midrule[1pt]
STM $^{\ast}$ \cite{STM_Oh_2019_ICCV}   & 81.8       & 79.2       & 84.3      & -          & -          & -          & 79.4  & 79.7  & 84.2  & 72.8  & 80.9 & -     & -     & -     & -    & -    \\
AFB-URR $^{\ast}$ \cite{AFBURR_NEURIPS2020_23483314}                  & 76.9       & 74.4       & 79.3      & -          & -          & -          & 79.6  & 78.8  & 83.1  & 74.1  & 82.6 & -     & -     & -     & -    & -    \\
CFBI $^{\ast}$ \cite{CFBI_10.1007/978-3-030-58558-7_20}   & 81.9       & 79.1       & 84.6      & 74.8       & 71.1       & 78.5       & 81.4  & 81.1  & 85.8  & 75.3  & 83.4 & 81.0    & 80.6  & 85.1  & 75.2 & 83   \\
RMNet $^{\ast}$ \cite{RMNet_Xie_2021_CVPR}                    & 83.5       & 81.0         & 86.0        & 75.0         & 71.9       & 78.1       & 81.5  & 82.1  & 85.7  & 75.7  & 82.4 & -     & -     & -     & -    & -    \\
SST \cite{SSTVOS_Duke_2021_CVPR}     & 82.5       & 79.9       & 85.1      & -          & -          & -          & 81.7  & 81.2  & -     & 76.0    & -    & 81.8  & 80.9  & -     & 76.6 & -    \\
MiVOS $^{\ast \dag}$ \cite{MiVOS_Cheng_2021_CVPR}   & 84.5       & 81.7       & 87.4      & 78.6       & 74.9       & 82.2       & 82.6  & 81.1  & 85.6  & 77.7  & 86.2 & 82.4  & 80.6  & 84.7  & 78.1 & 86.4 \\
HMMN $^{\ast}$ \cite{HMMN_Seong_2021_ICCV}   & 84.7       & 81.9       & 87.5      & 78.6       & 74.7       & 82.5       & 82.6  & 82.1  & 87.0    & 76.8  & 84.6 & 82.5  & 81.7  & 86.1  & 77.3 & 85.0   \\
JOINT \cite{JOINT_Mao_2021_ICCV}                    & 83.5       & 80.8       & 86.2      & -          & -          & -          & 83.1  & 81.5  & 85.9  & 78.7  & 86.5 & 82.8  & 80.8  & 84.8  & 79.0   & 86.6 \\
STCN $^{\ast}$ \cite{STCN_NEURIPS2021_61b4a64b}   & 85.4       & 82.2       & 88.6      & 76.1       & 72.7       & 79.6       & 83.0    & 81.9  & 86.5  & 77.9  & 85.7 & 82.7  & 81.1  & 85.4  & 78.2 & 85.9 \\
STCN $^{\ast \dag}$ \cite{STCN_NEURIPS2021_61b4a64b}   & 85.3       & 82.0         & 88.6      & 77.8       & 74.3       & 81.3       & 84.3  & 83.2  & 87.9  & 79.0    & 87.3 & 84.2  & 82.6  & 87.0    & 79.4 & 87.7 \\
R50-AOT-L $^{\ast}$ \cite{AOT_NEURIPS2021_147702db}                 & 84.9  & 82.3  & 87.5  & 79.6  & 75.9  & 83.3  & 85.5  & 84.5  & 89.5  & 79.6  & 88.2  & 85.3  & 83.9  & 88.8  & 79.9  & 88.5  \\
SwinB-AOT-L $^{\ast}$ \cite{AOT_NEURIPS2021_147702db}              & 85.4 & 82.4 & 88.4 & 81.2 & 77.3 & 85.1 & 85.1 & 85.1 & 90.1 & 78.4 & 86.9 & 85.3 & 84.6 & 89.5 & 79.3 & 87.7 \\
R50-DeAOT-L $^{\ast}$ \cite{DeAOT_yang2022decoupling} & 85.2 & 82.2 & 88.2 & 80.7 & 76.9 & 84.5 & 85.9 & 84.9 & 90.1 & 79.8 & 88.7 & 85.6 & 84.2 & 89.2 & 80.2 & 88.8 \\
SwinB-DeAOT-L $^{\ast}$ \cite{DeAOT_yang2022decoupling} & 86.2 & 83.1 & 89.2 & 82.8 & 78.9 & 86.7 & 86.3 & 85.4 & 90.7 & 80.1 & 89.0 & \textcolor{blue}{86.4} & \textcolor{black}{85.4} & \textcolor{blue}{90.3} & 80.5  & \textcolor{blue}{89.3} \\
XMem \cite{XMem_10.1007/978-3-031-19815-1_37}    & 84.5       & 81.4       & 87.6      & 79.8       & 76.3       & 83.4       & 84.3  & 83.9  & 88.8  & 77.7  & 86.7 & 84.2  & 83.8  & 88.3  & 78.1 & 86.7 \\
XMem $^{\ast}$  \cite{XMem_10.1007/978-3-031-19815-1_37}   & 86.2       & 82.9       & 89.5      & 81.0         & 77.4       & 84.5       & 85.7  & 84.6  & 89.3  & 80.2  & 88.7 & 85.5  & 84.3  & 88.6  & 80.3 & 88.6 \\
XMem $^{\ast \dag}$  \cite{XMem_10.1007/978-3-031-19815-1_37}   & 87.7       & 84.0         & 91.4      & 81.2       & 77.6       & 84.7       & 86.1  & 85.1  & 89.8  & 80.3  & \textcolor{blue}{89.2} & 85.8  & 84.8  & 89.2  & 80.3 & 88.8 \\
ISVOS $^{\ast}$  \cite{ISVOS_Wang2022LookBY}   & 87.1       & 83.7       & 90.5      & 82.8         & 79.3       & 86.2       & 86.3  & 85.5  & 90.2  & 80.5  & 88.8 & 86.1  & 85.2  & 89.7  & 80.7 & 88.9 \\
ISVOS $^{\ast \dag}$  \cite{ISVOS_Wang2022LookBY}   & \textcolor{black}{88.2}       & \textcolor{black}{84.5}         & \textcolor{black}{91.9}      & \textcolor{black}{84.0}       & \textcolor{black}{80.1}       & \textcolor{black}{87.8}       & \textcolor{blue}{86.7}  & \textcolor{blue}{86.1}  & \textcolor{blue}{90.8}  & \textcolor{blue}{81.0}  & \textcolor{black}{89.0} & 86.3  & 85.2  & 89.7  & \textcolor{blue}{81.0} & 89.1 \\
\textcolor{black}{SimVOS-B} $^{\ast}$  \cite{SimVOS}   & \textcolor{black}{88.0}       & \textcolor{black}{85.0}         & \textcolor{black}{91.0}      & \textcolor{black}{80.4}       & \textcolor{black}{76.1}       & \textcolor{black}{84.6}       & \textcolor{black}{-}  & \textcolor{black}{-}  & \textcolor{black}{-}  & \textcolor{black}{-}  & \textcolor{black}{-} & \textcolor{black}{84.2}  & \textcolor{black}{83.1}  & \textcolor{black}{-}  & \textcolor{black}{79.1} & \textcolor{black}{-} \\
\textcolor{black}{DEVA} $^{\ast}$  \cite{DEVA}   & \textcolor{black}{86.8}       & \textcolor{black}{83.6}         & \textcolor{black}{90.0}      & \textcolor{black}{82.3}       & \textcolor{black}{78.7}       & \textcolor{black}{85.9}       & \textcolor{black}{85.9}  & \textcolor{black}{85.5}  & \textcolor{black}{90.1}  & \textcolor{black}{79.7}  & \textcolor{black}{88.2} & \textcolor{black}{85.5}  & \textcolor{black}{85.0}  & \textcolor{black}{89.4}  & \textcolor{black}{79.7} & \textcolor{black}{88.0} \\
\textcolor{black}{Cutie} $^{\ast}$  \cite{Cutie}   & \textcolor{blue}{88.8}       & \textcolor{blue}{85.4}         & \textcolor{blue}{92.3}      & \textcolor{blue}{84.2}       & \textcolor{blue}{80.6}       & \textcolor{blue}{87.7}       & \textcolor{black}{86.1}  & \textcolor{black}{85.8}  & \textcolor{black}{90.5}  & \textcolor{black}{80.0}  & \textcolor{black}{88.0} & \textcolor{black}{86.1}  & \textcolor{blue}{85.5}  & \textcolor{black}{90.0}  & \textcolor{black}{80.6} & \textcolor{black}{88.3} \\
\midrule[1pt]
Ours              & 89.1 &	85.9 &	92.2 &	87.0 &	83.4 &	90.6 &	86.0 &	86.0 &	91.0 &	79.5 &	87.5 &	86.2 &	85.7  &	90.5 &	80.4 &	88.2 \\
Ours $^{\ast}$ & 89.7 &	86.7 & 92.7     & 87.6 & 84.2 & 91.1   & 87.0 & 86.2 & 91.0 & 81.4 & 89.3 & 87.0 & 86.1 & 90.6 & 82.0 & 89.5 \\
Ours $^{\ast \dag}$ & \textcolor{red}{90.1} & \textcolor{red}{87.0} & \textcolor{red}{93.2}     & \textcolor{red}{88.1} & \textcolor{red}{84.7} & \textcolor{red}{91.6}   & \textcolor{red}{87.6}  & \textcolor{red}{86.4}  & \textcolor{red}{91.0}  & \textcolor{red}{82.2}  & \textcolor{red}{90.7} & \textcolor{red}{87.4}  & \textcolor{red}{86.5}  & \textcolor{red}{90.9}  & \textcolor{red}{82.0} & \textcolor{red}{90.3} \\
\bottomrule[1.5pt]
\end{tabular}
\end{center}
\label{tab:sota_D17_YTB}
\end{table*}

\subsection{Training Objective}

For each training sample sequence, we sample $T$ frames which contain one annotated frame and $T-1$ frames need to be predicted \textcolor{black}{and optimized}. \textcolor{black}{In order to simulate the memory bank during the inference process, in addition to the first frame, subsequent frames also make use of the previously generated predictions as a reference.}
We separate and parallelize multiple targets, and cast the video object segmentation task as a classification task that classifies each pixel into targets or \textcolor{black}{the} background.
To optimize our model, we follow~\cite{AOT_NEURIPS2021_147702db, XMem_10.1007/978-3-031-19815-1_37} to adopt the combination of bootstrapped cross entropy loss~\cite{DBLP:journals/corr/ReedLASER14} and dice loss~\cite{DBLP:conf/acl/LiSMLWL20,DBLP:conf/3dim/MilletariNA16} as our optimization objective. Specifically, \textcolor{black}{assuming that the maximum number of targets appearing during training is $N_{obj}$}, we denote the predicted mask of each frame as $\hat{\boldsymbol{M}_{t}} \in \mathbb{R}^{(1+N_{obj}) \times HW}$ and the ground-truth as $\boldsymbol{M}_{t} \in \mathbb{R}^{(1+N_{obj}) \times HW}$. The loss function is defined as:

\begin{equation}
\mathcal{L} = \sum_{t=2}^{T} \mathcal{L}_{BCE}(\boldsymbol{M}_{t}, \hat{\boldsymbol{M}_{t}}) + \mathcal{L}_{Dice}(\boldsymbol{M}_{t}, \hat{\boldsymbol{M}_{t}}),
\label{eq:loss}
\end{equation}
where $T$ is the length of the sequence, and $\mathcal{L}_{BCE}$ and $\mathcal{L}_{Dice}$ are the bootstrapped cross entropy loss and dice loss, respectively. Since the first frame is the given annotations, we start calculating the loss in the second frame.

\section{Experiments}

\subsection{Implementation Details}

\noindent \textcolor{black}{\textbf{Model.}}
We apply ConvMAE-base~\cite{ConvMAE_gao2022convmae}, a \textcolor{black}{hierarchical Vision Transformer} network that replaces \textcolor{black}{the patch embedding layer of ViT\cite{vit}} with fewer lightweight CNN blocks \textcolor{black}{to provide multi-scale feature}, as our backbone with its MAE pre-training~\cite{MAE_He_2022_CVPR}.
\textcolor{black}{Beyond the backbone, we} replicate the last two blocks of the pre-trained backbone to \textcolor{black}{update} compressed memory with decoder tokens and copy the last block to \textcolor{black}{\emph{enhance}} the current tokens \textcolor{black}{with compressed memory}.
\textcolor{black}{The detailed implementation of the decoder is} shown in Fig.~\ref{fig:arch_decoder}, we first reshape the current tokens and enhanced current tokens \textcolor{black}{into 2D shape, concatenate them along channel dimension, and fuse them. Then the fused features are iteratively upsampled} by 2$\times$ at a time until $1/4$ scale while \textcolor{black}{fused} with the skip connections from the \textcolor{black}{hierarchical} backbone.
Finally, we generate the single-channel logit with a 3 $\times$ 3 convolution and bilinearly upsample it to the input resolution.
In addition, we fuse and concatenate the internal features in $1/16$, $1/8$, and $1/4$ scale within the decoder to generate the \textbf{decoder token} \textcolor{black}{for further updating the compressed token}.
\textcolor{black}{In the multi-object scenario, we first separate multiple targets and process them in parallel in our network to generate the mask prediction for each target $\hat{\boldsymbol{m}_{t, i}} \in \mathbb{R}^{1 \times HW}$ which means we classify each pixel into foreground or background, \ie, target region or non-target region.
After that, we apply the soft-aggregation operation~\cite{STM_Oh_2019_ICCV} to merge the predictions $\hat{\boldsymbol{M}_{t}} \in \mathbb{R}^{(1+N_{obj}) \times HW}$ from different $N_{obj}$ targets as:
}

\color{black}
\vspace{-3mm}
\begin{equation}
\begin{aligned}
    \hat{\boldsymbol{m}}_{t, targets} &= \mathrm{Concate}_{i=1}^{N_{obj}}[\hat{\boldsymbol{m}_{t, i}}], \\
    \hat{\boldsymbol{m}}_{t, background} &= 1 - \prod_{i=1}^{N_{obj}}\hat{\boldsymbol{m}_{t, i}}, \\
    \hat{\boldsymbol{m}}_{t} &= \mathrm{Concate}[\hat{\boldsymbol{m}}_{t, background}; \hat{\boldsymbol{m}}_{t, targets}], \\
    \hat{\boldsymbol{M}}_{t} &= \mathrm{Softmax}(\log(\hat{\boldsymbol{m}}_{t})). \\
\end{aligned}
\label{eq:soft-aggregation}
\end{equation}
\color{black}

\noindent \textbf{Training.} 
Following~\cite{STM_Oh_2019_ICCV, KMN_10.1007/978-3-030-58542-6_38, STCN_NEURIPS2021_61b4a64b, AOT_NEURIPS2021_147702db, XMem_10.1007/978-3-031-19815-1_37}, we first pre-train our network on synthetic video sequences generated from static image datasets~\cite{CSSD_7182346, Wang_2017_CVPR, Li_2020_CVPR, Zeng_2019_ICCV, Cheng_2020_CVPR}.
Then, we perform the main-training on DAVIS 2017~\cite{DAVIS17_Pont-Tuset_arXiv_2017} and YouTube-VOS 2019~\cite{youtubeVOS_xu2018youtube} datasets with curriculum sampling \textcolor{black}{strategy}~\cite{STM_Oh_2019_ICCV}.
\textcolor{black}{In addition,} we also provide results that are \textcolor{black}{advanced} pre-trained on BL30K~\cite{MiVOS_Cheng_2021_CVPR} optionally to further improve the performance. The models pre-trained with small static image datasets and big BL30K dataset are denoted by $^{\ast}$ and $^{\dag}$.
To improve the robustness of compressed memory in temporal sequence, we set it to accept the update of reference tokens with a probability of 0.6 for each frame.
For each training sample sequence, we set the maximum number $N_{obj}$ of randomly selected targets to three.
\textcolor{black}{The crop sizes of images are $384 \times 384$.}
The sequence lengths $T$ of pre-training and main training are three and four respectively, and a maximum of two past frames are randomly selected to be reference frames.

\begin{table}[]
\small
\fontsize{9pt}{4mm}\selectfont
\setlength{\tabcolsep}{2.8mm}
\caption{Quantitative evaluation on DAVIS 2016 val split.}
\begin{center}
\begin{tabular}{l|ccc}
\toprule[1.5pt]
Method & $\mathcal{J}\&\mathcal{F}$       & $\mathcal{J}$          & $\mathcal{F}$ \\ 
\midrule[1pt]
STM $^{\ast}$ \cite{STM_Oh_2019_ICCV}                       & 89.3       & 88.7      & 89.9      \\
CFBI $^{\ast}$ \cite{CFBI_10.1007/978-3-030-58558-7_20}                    & 89.4       & 88.3      & 90.5      \\
CFBI+ $^{\ast}$ \cite{CFBIP_yang2020}                    & 89.9       & 88.7      & 91.1      \\ 
RMNet $^{\ast}$  \cite{RMNet_Xie_2021_CVPR}                   & 88.8       & 88.9      & 88.7      \\
KMN $^{\ast}$ \cite{KMN_10.1007/978-3-030-58542-6_38}                    & 90.5       & 89.5      & 91.5      \\ 
MiVOS $^{\ast \dag}$ \cite{MiVOS_Cheng_2021_CVPR}                     & 91.0       & 89.6      & 92.4      \\
HMMN $^{\ast}$ \cite{HMMN_Seong_2021_ICCV}                      & 90.8       & 89.6      & 92.0        \\
RDE $^{\ast}$ \cite{RDE_DBLP:conf/cvpr/LiHXZP022}                      & 91.1       & 89.7      & 92.5      \\
STCN $^{\ast}$ \cite{STCN_NEURIPS2021_61b4a64b}                      & 91.6       & 90.8      & 92.5      \\
STCN $^{\ast \dag}$ \cite{STCN_NEURIPS2021_61b4a64b}                       & 91.7       & 90.4      & 93.0        \\
R50-AOT-L $^{\ast}$ \cite{AOT_NEURIPS2021_147702db}                & 91.1       & 90.1      & 92.1      \\
SwinB-AOT-L $^{\ast}$ \cite{AOT_NEURIPS2021_147702db}              & 92.0       & 90.7      & 93.3      \\
R50-DeAOT-L $^{\ast}$ \cite{DeAOT_yang2022decoupling}              & 92.3       & 90.5      & 94.0        \\
SwinB-DeAOT-L $^{\ast}$ \cite{DeAOT_yang2022decoupling}            & 92.9       & 91.1      & 94.7      \\
XMem \cite{XMem_10.1007/978-3-031-19815-1_37}                     & 90.8       & 89.6      & 91.9      \\
XMem $^{\ast}$ \cite{XMem_10.1007/978-3-031-19815-1_37}                     & 91.5       & 90.4      & 92.7      \\
XMem $^{\ast \dag}$ \cite{XMem_10.1007/978-3-031-19815-1_37}                     & 92.0       & 90.7      & 93.2      \\
ISVOS $^{\ast}$ \cite{ISVOS_Wang2022LookBY}                     & 92.6       & 91.5      & 93.7      \\
ISVOS $^{\ast \dag}$ \cite{ISVOS_Wang2022LookBY}                     & 92.8       & 91.8      & 93.8      \\
\midrule[1pt]
Ours $^{\ast}$                    & 92.1 & 90.6 & 93.6     \\
Ours $^{\ast \dag}$                    & 92.4       & 90.4      & 94.4     \\
\bottomrule[1.5pt]
\end{tabular}

\end{center}
\label{table:sota_D16}
\vspace{-5mm}
\end{table}

\begin{figure*}[pt]
\centering
\includegraphics[width=1.0\linewidth]{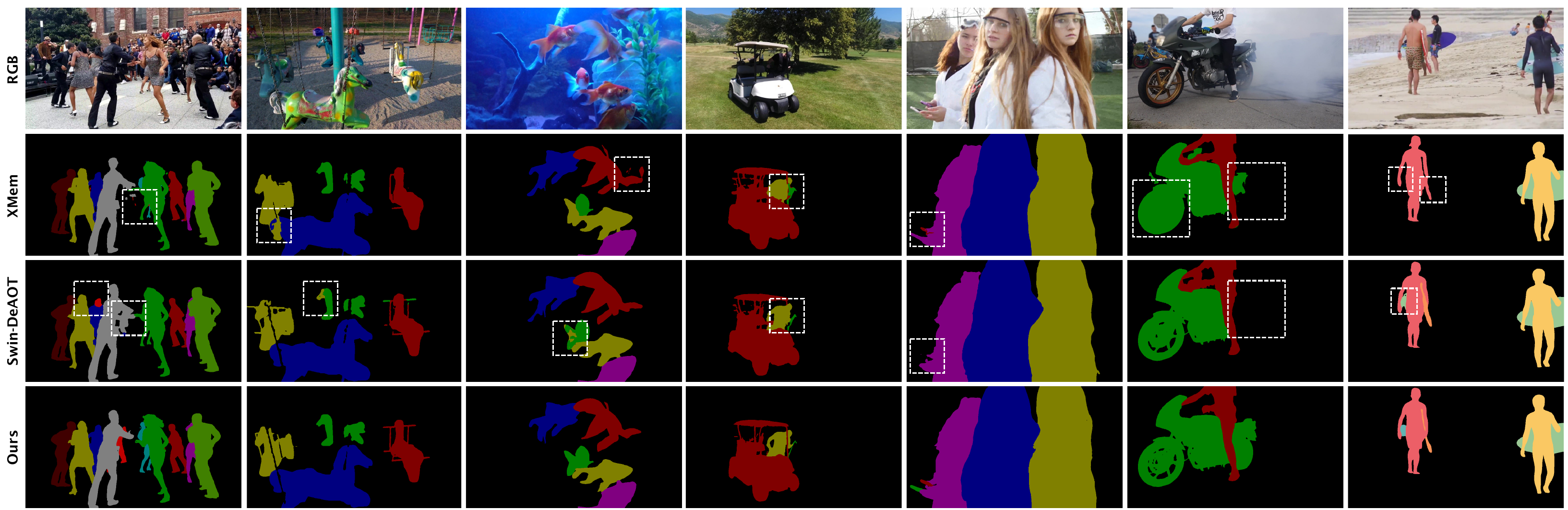}
\caption{Qualitative comparisons of our model with XMem~\cite{XMem_10.1007/978-3-031-19815-1_37} and Swin-DeAOT~\cite{DeAOT_yang2022decoupling} on DAVIS and YouTube-VOS benchmark.
The difficulty in the first three columns lies in recognizing similar objects, while the remaining columns lie in tiny targets and object boundaries.
We mark their failures in the white dashed boxes.
Our model outperforms them in terms of detailing and discriminating similarities.}
\vspace{-3mm}
\label{fig:results_fig0}
\end{figure*}

For optimization, we use the AdamW~\cite{Adam_DBLP:journals/corr/KingmaB14, loshchilov2018decoupled} optimizer with a learning rate of 5e-5 and a weight decay of 0.05.
To avoid over-fitting, the initial learning rate of the backbone is reduced to 0.1 of other network parts.
We pre-train our model for 150K iterations with batch size 24 on static image datasets, for 300K iterations with batch size 12 on BL30K dataset, and the main-training \textcolor{black}{stage} lasts 160K iterations with batch size 12.
We drop the learning rate by a factor of 10 after the first 200K and 100K iterations on BL30K pre-training and VOS main-training, but not drop on static image pre-traing.

\noindent \textbf{Inference.}
\textcolor{black}{As for the memory bank, we} simply store the reference frames with their masks ground-truth or prediction.
For the shorter DAVIS dataset, we only keep the first and the previous frames.
For the longer YouTube-VOS dataset, we simply implement a first-in-first-out queue by memorizing every 5th frame following previous work~\cite{STM_Oh_2019_ICCV, STCN_NEURIPS2021_61b4a64b} and set the maximum frame size \textcolor{black}{of the queue} to 3 in addition to the previous frames for a diversity of target information.

For DAVIS evaluation, we use the 480p 24FPS videos, while for YouTube-VOS, we use all input frames and please note that we re-run AOT~\cite{AOT_NEURIPS2021_147702db} and DeAOT~\cite{DeAOT_yang2022decoupling} with the same all input frame for fair comparison.
We resize the videos to 480p and interpolate the positional embedding to the input size \textcolor{black}{which is inconsistent with training size}.
We adopt top-K \textcolor{black}{filter~\cite{MiVOS_Cheng_2021_CVPR} for current tokens} in all Joint Modeling Blocks by default.
Specifically, the current tokens \textcolor{black}{interact with}
themselves and all reference tokens \textcolor{black}{to obtain} the attention matrix $A \in \mathbb{R}^{N_{x} \times (N_{x}+N_{z})}$, and then keep all of current tokens and the top-K of $N_{z}$ reference tokens to \textcolor{black}{modify the attention matrix as} $A' \in \mathbb{R}^{N_{x} \times (N_{x}+N_{topK})}$ before \emph{softmax} to perform robust matching while maintaining the details in the current frame.

\subsection{Comparison with the State-of-the-art Methods}

\subsubsection{Mainstream Benchmark}

\noindent \textbf{Datasets and evaluation metrics.}
We report the results on DAVIS 2016/2017~\cite{DAVIS16_Perazzi2016, DAVIS17_Pont-Tuset_arXiv_2017} and YouTube-VOS 2018/2019~\cite{youtubeVOS_xu2018youtube} \textcolor{black}{benchmarks}.
The evaluation metrics include region similarity $\mathcal{J}$, contour accuracy $\mathcal{F}$, and their average $\mathcal{J}\&\mathcal{F}$, and \textcolor{black}{it} also reports \textcolor{black}{results that belong} the seen and unseen categories for YouTube-VOS dataset. We evaluate all the results  \textcolor{black}{with official tools for DAVIS or on official evaluation servers for YouTube-VOS}.

\noindent \textbf{DAVIS 2017.}
DAVIS 2017~\cite{DAVIS17_Pont-Tuset_arXiv_2017} is a multiple objects extension of DAVIS 2016, whose validation split has 30 videos with 59 objects, and the test split contains 30 more challenging videos with 89 objects.
Table~\ref{tab:sota_D17_YTB} shows our model significantly outperforms all existing works both on the validation (\textbf{89.7\%}) and test-dev (\textbf{87.6\%}) split.
Remarkably, compared to the SwinB-DeAOTL \textcolor{black}{and Cutie, the previous state-of-the-art methods, our model even outperforms them by a large margin \textbf{without} any synthetic pre-training}.
The excellent performance of the \textcolor{black}{more} difficult, complex test split demonstrates that our model works great in handling both complex and detailed targets.

\begin{table}[]
\small
\fontsize{9pt}{4mm}\selectfont
\setlength{\tabcolsep}{0.7mm}
\caption{Multi-scale evaluation on DAVIS and YouTube-VOS.}
\begin{center}
\begin{subtable}{}
    \begin{tabular}{l|ccc|ccc|ccc}
\toprule[1.5pt]
\multirow{2}{*}{Method} & \multicolumn{3}{c|}{DAVIS 2016 val} & \multicolumn{3}{c|}{DAVIS 2017 val} & \multicolumn{3}{c}{DAVIS 2017 test} \\ \cline{2-10} 
\rule{0pt}{9pt} & $\mathcal{J}\&\mathcal{F}$       & $\mathcal{J}$          & $\mathcal{F}$ & $\mathcal{J}\&\mathcal{F}$       & $\mathcal{J}$          & $\mathcal{F}$ & $\mathcal{J}\&\mathcal{F}$       & $\mathcal{J}$          & $\mathcal{F}$ \\
\midrule[1pt]
\textcolor{black}{XMem $^{\ast}$}  & - & - & -   & 86.2 &	82.9 & 89.5 & 81.0 & 77.4 & 84.5   \\
\textcolor{black}{XMem $^{\ast}_{MS}$}   & - & - & - & 88.2 & 85.4 & 91.0 & 83.1 & 79.7 & 86.4 \\
\textcolor{black}{ISVOS $^{\ast}$}  & - & - & -   & 87.1 & 83.7 & 90.5 & - & - & -   \\
\textcolor{black}{ISVOS $^{\ast}_{MS}$}   & - & - & - & 88.6 & 85.8 & 91.4 & - & - & - \\
Ours $^{\ast}$  & 92.1 & 90.6 & 93.6   & 89.7 &	86.7 & 92.7 & 87.6 & 84.2 & 91.1   \\
Ours $^{\ast \dag}$   & 92.4 & 90.4 & 94.4   & 90.1 & 87.0 & 93.2     & 88.1 & 84.7 & 91.6 \\
Ours $^{\ast}_{MS}$   & 92.3 & 91.2 & 93.4 & 90.0 & 87.2 & 92.9 & 88.1 & 84.8 & 91.3 \\
Ours $^{\ast \dag}_{MS}$   & 92.8 & 91.3 & 94.4 & 90.6 & 87.8 & 93.5 & 88.5 & 85.1 & 91.8 \\
\bottomrule[1.5pt]
\end{tabular}

    \label{subfig:A}
    \end{subtable}%
    \hspace{0.1\textwidth}
    \vspace{10pt}
    
    \begin{subtable}{}
    \begin{tabular}{l|ccccc|ccccc}
\toprule[1.5pt]
\multirow{2}{*}{Method} & \multicolumn{5}{c|}{YouTube-VOS 2018 val} & \multicolumn{5}{c}{YouTube-VOS 2019 val} \\ \cline{2-11} 
\rule{0pt}{9pt} & $\mathcal{G}$     & $\mathcal{J}_S$    & $\mathcal{F}_S$    & $\mathcal{J}_U$    & $\mathcal{F}_U$   & $\mathcal{G}$     & $\mathcal{J}_S$    & $\mathcal{F}_S$    & $\mathcal{J}_U$   & $\mathcal{F}_U$     \\
\midrule[1pt]
\textcolor{black}{STCN $^{\ast}$}   & - & - & - & - & - & 84.2 & 82.6 & 87.0 & 79.4 & 87.7  \\
\textcolor{black}{STCN $^{\ast}_{MS}$}   & - & - & - & - & - & 85.2 & 83.5 & 87.8 & 80.7 & 88.7 \\
\textcolor{black}{XMem $^{\ast}$}   & 85.7 & 84.6 & 89.3 & 80.2 & 88.7 & 85.5 & 84.3 & 88.6 & 80.3 & 88.6  \\
\textcolor{black}{XMem $^{\ast}_{MS}$}   & 86.7   & 85.3   & 89.9   & 81.7   & 89.9   & 86.4   & 84.9   & 89.2   & 81.8   & 89.8 \\
Ours $^{\ast}$   & 87.0 & 86.2 & 91.0 & 81.4 & 89.3 & 87.0 & 86.1 & 90.6 & 82.0 & 89.5  \\
Ours $^{\ast \dag}$   & 87.6  & 86.4  & 91.0  & 82.2  & 90.7 & 87.4  & 86.5  & 90.9  & 82.0 & 90.3 \\
Ours $^{\ast}_{MS}$   & 87.2   & 86.5   & 91.3   & 81.8   & 89.4   & 87.4   & 86.4   & 91.0   & 82.4   & 89.8 \\
Ours $^{\ast \dag}_{MS}$   & 87.6   & 86.4   & 91.0   & 82.3   & 90.7   & 87.5   & 86.5   & 90.9   & 82.2   & 90.4 \\ 
\bottomrule[1.5pt]
\end{tabular}

    \label{subfig:B}
    \end{subtable}
\end{center}
\label{table:MS_test}
\end{table}

\noindent \textbf{YouTube-VOS.}
YouTube-VOS~\cite{youtubeVOS_xu2018youtube} is the latest large-scale benchmark for multi-object video segmentation which is about 37 times larger than DAVIS 2017. It has 3471 videos in the training split with 65 categories and 474/507 videos in the validation 2018/2019 split which contain 26 unseen categories that do not exist in the training split to evaluate the generalization ability of algorithms.
As shown in Table~\ref{tab:sota_D17_YTB}, our model achieves superior performance both on YouTube-VOS 2018 (\textbf{87.0\%}) and 2019 (\textbf{87.0\%}). With the additional larger BL30K pre-training, our model further improves the performance on 2018 (\textbf{87.6\%}) and 2019 (\textbf{87.4\%}).

\noindent \textbf{DAVIS 2016.}
DAVIS 2016~\cite{DAVIS16_Perazzi2016} is a single-object benchmark containing 20 videos in the validation split.
As shown in Table~\ref{table:sota_D16}, our network achieves competitive performance (\textbf{92.1\%}) with the previous state-of-the-art methods.

\noindent \textbf{Qualitative Results.}
We compare our method qualitatively with XMem~\cite{XMem_10.1007/978-3-031-19815-1_37} and Swin-DeAOT~\cite{DeAOT_yang2022decoupling} in Fig.~\ref{fig:results_fig0}, all three models are pre-trained with image dataset \textcolor{black}{and main-trained on DAVIS 2017 and Youtube-VOS 2017 datasets}.
The first three columns show that our model greatly improves the discrimination of similar objects compared to previous methods, demonstrating the importance of joint modeling and our compressed memory.
The remaining columns show that our model handles tiny targets and object boundaries as well, suggesting that joint modeling at low levels handles details better, \textcolor{black}{which} are difficult to capture in previous post-matching methods.

\noindent \textbf{Multi-scale evaluation.}
To further improve the performance without modifying the model and training strategy, we employ multi-scale evaluation which is a general trick used in segmentation tasks.
Specifically, we adopt scale variation and vertical mirroring and process the inputs at different scales independently, and finally average \textcolor{black}{these} output probability maps.
We adopt multiple scales \{1.0, 1.1, 1.2\} $\times$ 480p on DAVIS and \{1.0, 1.1\} $\times$ 480p on YouTube-VOS \textcolor{black}{without changing other inference settings.}
Evaluation results are shown in Table~\ref{table:MS_test}, indicating that our model can further improve performance through multi-scale evaluation.

\subsubsection{New Benchmark}

To demonstrate the generality and excellent performance of our approach, we test our method on several recently proposed benchmarks: MOSE~\cite{MOSE_DBLP:journals/corr/abs-2302-01872}, VISOR~\cite{VISOR_DBLP:conf/nips/DarkhalilSZMKHF22}, VOST~\cite{VOST_DBLP:conf/cvpr/TokmakovLG23}, and LVOS~\cite{LVOS_hong2022lvos}.
To show more clearly the performance of the model on these new benchmarks, we compare ours to several methods, including XMem~\cite{XMem_10.1007/978-3-031-19815-1_37}, Swinb-AOTL~\cite{AOT_NEURIPS2021_147702db}, Swinb-DeAOTL~\cite{DeAOT_yang2022decoupling}, \textcolor{black}{and Cutie~\cite{Cutie}} with their open-source code\footnote{\url{https://github.com/hkchengrex/XMem}} \footnote{\url{https://github.com/yoxu515/aot-benchmark}} \textcolor{black}{or papers}.

\begin{table}[]
\small
\fontsize{9pt}{4mm}\selectfont
\setlength{\tabcolsep}{3mm}
\caption{
Quantitative evaluation on MOSE validation split. \\
\textcolor{black}{The default training setting is to train on the training split of the MOSE dataset for the main training phase.}
}
\begin{center}
\begin{tabular}{l|ccc}
\toprule[1.5pt]
Method & $\mathcal{J}\&\mathcal{F}$       & $\mathcal{J}$          & $\mathcal{F}$ \\ 
\midrule[1pt]
SwinB-AOT-L \cite{AOT_NEURIPS2021_147702db}       & 62.3     & 57.8         & 66.8 \\
SwinB-AOT-L $^{\ast}$ \cite{AOT_NEURIPS2021_147702db}       & 64.6     & 60.2         & 69.0 \\
SwinB-DeAOT-L \cite{DeAOT_yang2022decoupling}     & 65.1     & 60.9         & 69.4 \\
SwinB-DeAOT-L $^{\ast}$ \cite{DeAOT_yang2022decoupling}     & \textcolor{black}{67.3}     & \textcolor{black}{62.9}         & \textcolor{black}{71.6} \\
XMem \cite{XMem_10.1007/978-3-031-19815-1_37}              & 56.3     & 51.8         & 60.8 \\
XMem $^{\ast}$  \cite{XMem_10.1007/978-3-031-19815-1_37}              & 59.0     & 54.6         & 63.4 \\
XMem $^{\ast \dag}$  \cite{XMem_10.1007/978-3-031-19815-1_37}              & 62.9     & 58.6         & 67.2 \\
\textcolor{gray}{DEVA} $^{\textcolor{gray}{\ast, +D17, +Y19, -MOSE}}$~\cite{DEVA}             & \textcolor{gray}{60.0}     & \textcolor{gray}{55.8}         & \textcolor{gray}{64.3} \\
\textcolor{gray}{DEVA} $^{\textcolor{gray}{\ast, +D17, +Y19}}$~\cite{DEVA}             & \textcolor{gray}{66.0}     & \textcolor{gray}{61.8}         & \textcolor{gray}{70.3} \\
\textcolor{gray}{Cutie} $^{\textcolor{gray}{\ast, +D17, +Y19, -MOSE}}$~\cite{Cutie}             & \textcolor{gray}{64.0}     & \textcolor{gray}{60.0}         & \textcolor{gray}{67.9} \\
\textcolor{gray}{Cutie} $^{\textcolor{gray}{\ast, +D17, +Y19}}$~\cite{Cutie}             & \textcolor{gray}{68.3}     & \textcolor{gray}{64.2}         & \textcolor{gray}{72.3} \\
\textcolor{black}{Cutie} $^{\ast}$ ~\cite{Cutie}             & \textcolor{blue}{69.2}     & \textcolor{blue}{65.2}         & \textcolor{blue}{73.3} \\
\midrule[1pt]
Ours              & 66.2     & 62.3         & 70.1 \\
Ours $^{\ast}$             & 69.7     & 65.8         & 73.6 \\
Ours $^{\ast \dag}$             & \textcolor{red}{70.2}     & \textcolor{red}{66.3}         & \textcolor{red}{74.0} \\
\bottomrule[1.5pt]
\end{tabular}

\end{center}
\label{table:sota_MOSE}
\end{table}

\begin{table}[]
\small
\fontsize{9pt}{4mm}\selectfont
\setlength{\tabcolsep}{1.1mm}
\caption{Quantitative evaluation on VISOR validation split.}
\begin{center}
\begin{tabular}{l|ccc|ccc}
\toprule[1.5pt]
\multirow{2}{*}{Method} & \multicolumn{3}{c|}{VISOR val} & \multicolumn{3}{c}{VISOR $\text{val unseen}$} \\ \cline{2-7} 
  & $\mathcal{J}\&\mathcal{F}$       & $\mathcal{J}$          & $\mathcal{F}$     & $\mathcal{J}\&\mathcal{F}$       & $\mathcal{J}$          & $\mathcal{F}$     \\
\midrule[1pt]
SwinB-AOT-L \cite{AOT_NEURIPS2021_147702db}       & 83.1     & 80.6         & 85.6         & 81.7     & 79.7         & 83.7        \\
SwinB-AOT-L $^{\ast}$ \cite{AOT_NEURIPS2021_147702db}       & 83.7     & 81.2         & 86.2         & 83.6     & 81.7         & 85.5         \\
SwinB-DeAOT-L \cite{DeAOT_yang2022decoupling}     & 84.4     & 81.8         & 87.1         & 83.8     & 81.8         & 85.8         \\
SwinB-DeAOT-L $^{\ast}$ \cite{DeAOT_yang2022decoupling}     & \textcolor{blue}{85.0}     & \textcolor{blue}{82.4}         & \textcolor{blue}{87.6}         & \textcolor{blue}{84.6}     & \textcolor{blue}{82.6}         & \textcolor{blue}{86.6}        \\
XMem   \cite{XMem_10.1007/978-3-031-19815-1_37}              & 66.4     & 64.2         & 68.6         & 60.6     & 58.1         & 63.0        \\
XMem  $^{\ast}$  \cite{XMem_10.1007/978-3-031-19815-1_37}              & 71.4     & 69.2         & 73.7         & 66.6     & 64.5         & 68.7         \\
XMem  $^{\ast \dag}$  \cite{XMem_10.1007/978-3-031-19815-1_37}              & 72.1     & 70.0        & 74.1         & 67.3     & 65.4         & 69.1         \\
\midrule[1pt]
Ours              & 85.1     & 82.9         & 87.2         & 84.6     & 82.9         & 86.3         \\
Ours  $^{\ast}$              & \textcolor{red}{85.9}     & \textcolor{red}{83.7}         & \textcolor{red}{88.0}         & \textcolor{red}{85.4}     & \textcolor{red}{83.7}         & \textcolor{red}{87.0}         \\
\bottomrule[1.5pt]
\end{tabular}
\end{center}
\label{table:sota_VISOR}
\end{table}

\begin{figure}[t]
\centering
\includegraphics[width=1.0\linewidth]{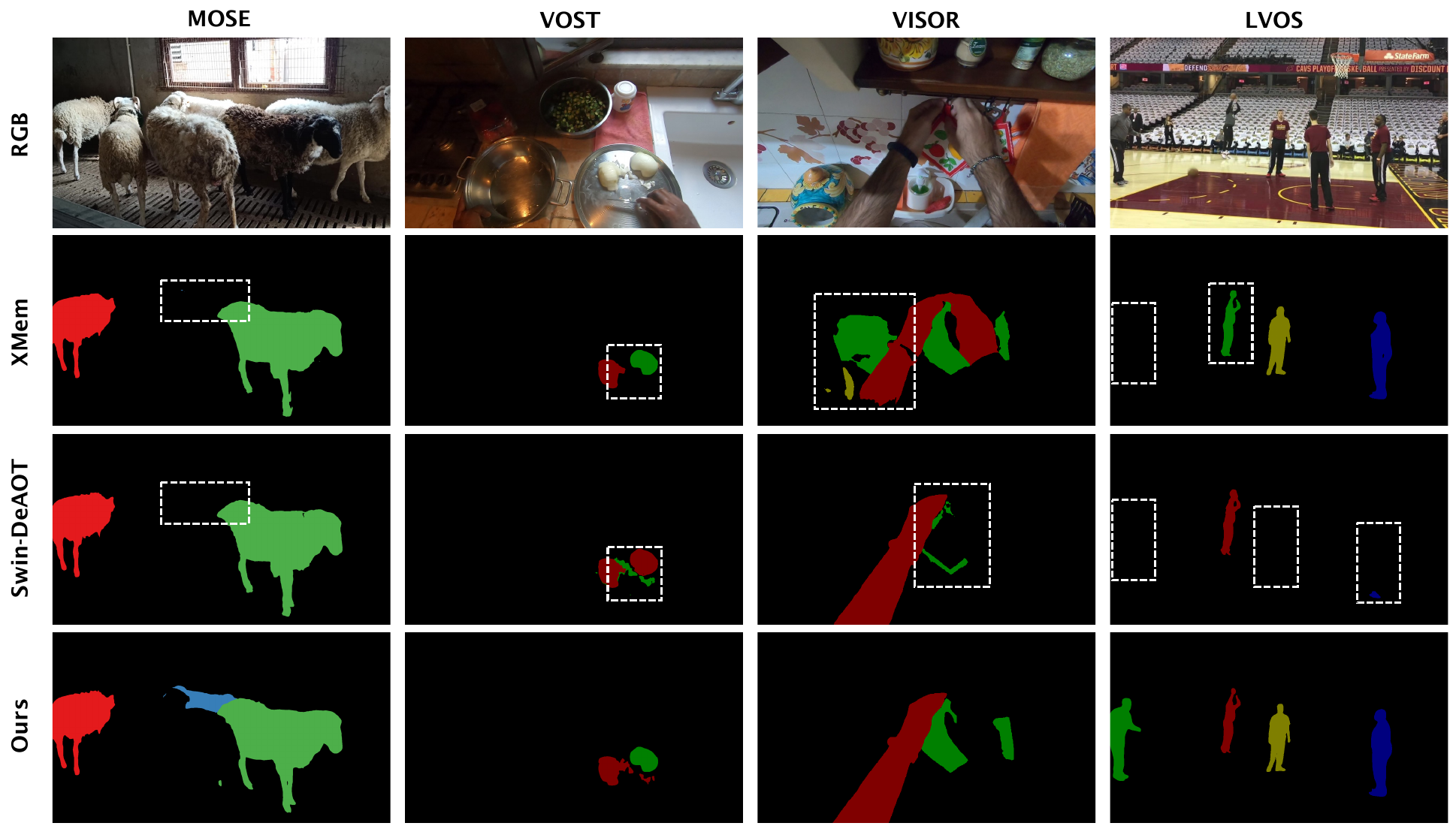}
\caption{Qualitative comparisons of our model with XMem~\cite{XMem_10.1007/978-3-031-19815-1_37} and Swin-DeAOT~\cite{DeAOT_yang2022decoupling} on four new benchmark: MOSE~\cite{MOSE_DBLP:journals/corr/abs-2302-01872}, VOST~\cite{VOST_DBLP:conf/cvpr/TokmakovLG23}, VISOR~\cite{VISOR_DBLP:conf/nips/DarkhalilSZMKHF22}, and LVOS~\cite{LVOS_hong2022lvos}.
The \textcolor{black}{various} difficulties include congestion, large transformation, egocentric, and long-term tracking respectively.
}
\label{fig:results_fig_MOSE_VOST}
\end{figure}

\noindent \textbf{Training and Inference.}
We follow the setting of each paper to train on the train split of each dataset.
MOSE~\cite{MOSE_DBLP:journals/corr/abs-2302-01872} and VISOR~\cite{VISOR_DBLP:conf/nips/DarkhalilSZMKHF22} benchmarks follow the regular setting, we train models on their train split with the same settings as main-training.
For the VOST benchmark~\cite{VOST_DBLP:conf/cvpr/TokmakovLG23}, we fine-tune 40K iterations on it after main-training.
As for the LVOS benchmark~\cite{LVOS_hong2022lvos}, we provide two settings: the first one is to directly use the checkpoint trained on DAVIS and the YouTube-VOS dataset to show the zero-shot performance, and the other one is to fine-tune 10K iterations on LVOS train split (\textcolor{black}{the fewer iterations due to its smaller training splits which prevents overfitting}).
The models pre-trained with additional \textcolor{black}{image} datasets are also denoted with $^{\ast}$ and $^{\dag}$.
Note that to illustrate the generalization and robustness of the model, the training parameters of each model in the three benchmarks are consistent with those in main-training without any changes.

During inference, we also adopt a simple first-in-first-out queue by memorizing every 5th frame.
Based on the different lengths of the dataset, we use only the first and previous frames on the shorter VISOR dataset (consistent with DAVIS) and set the maximum frame size to 3 in addition to the previous frames on the longer MOSE, VOST, and LVOS dataset (consistent with YouTube-VOS).
We resize videos to 480p and apply top-K filtering on all blocks.

\noindent \textbf{Evaluation metrics.}
Similar to DAVIS \textcolor{black}{and} YouTube-VOS, the evaluation metrics also include region similarity $\mathcal{J}$, contour accuracy $\mathcal{F}$, and their average $\mathcal{J}\&\mathcal{F}$.
On the VISOR benchmark, we specifically report metrics of the unseen categories in the validation split \textcolor{black}{that are not present in the training set}.
On the VOST benchmark, we additionally report region similarity $\mathcal{J}_{tr}$ which represents the last 25\% of the frames in a sequence after the \textcolor{black}{transformations are} mostly \textcolor{black}{completed}, and calculate the gap $\Delta_{tr}$ between \textcolor{black}{them} and the whole video to show the robustness of the model to object transformations.
On the LVOS benchmark, we additionally report the standard deviation $\mathcal{V}$ which measures the fluctuation of accuracy to represent the temporal stability of models (the lower is better).
We evaluate all the results with official evaluation servers or official tools.

\noindent \textbf{MOSE.} MOSE~\cite{MOSE_DBLP:journals/corr/abs-2302-01872} is a new dataset for \textcolor{black}{complex scenes} with crowded and occluded objects.
It is most notably characterized by complex environments, including object disappearance-appearance, small/unobtrusive objects, heavy occlusion, and crowded environments. \textcolor{black}{Meanwhile,} it needs stronger association correlation, discriminatory, and long-term capabilities.
MOSE has 1507 videos in the training split and 311 videos in the valid split with 5,200 objects from 36 categories.
As shown in Table~\ref{table:sota_MOSE}, our model outperforms the \textcolor{black}{four} SOTA models on the validation split (\textbf{69.7\%}) with the synthetic image pre-training and improves performance with larger BL30K dataset pre-training (\textbf{70.2\%}).
The excellent performance shows that our joint modeling and compressed memory can handle the difficulties of complex scenes well such as crowded, occluded, and so on.

\noindent \textbf{VISOR.}
VISOR~\cite{VISOR_DBLP:conf/nips/DarkhalilSZMKHF22} is a new benchmark suite for segmenting hands and active objects in \textcolor{black}{egocentric videos}. 
It requires models that ensure short- and long-term consistency of pixel-level annotations, as objects undergo \textcolor{black}{transformations of interactions}.
VISOR has 5309 videos in the training split and 1251 videos in the validation split which contains 159 videos of unseen kitchens that have not appeared in the training split.
Table~\ref{table:sota_VISOR} shows our model outperforms three works on the validation, indicating that our model can be well adapted to egocentric videos, and exhibits favorable generalization on the unseen part of validation.

\begin{table}[]
\small
\fontsize{7pt}{4mm}\selectfont
\setlength{\tabcolsep}{1mm}
\caption{Quantitative evaluation on VOST validation and test split. \\
\textcolor{black}{XMem $^{\ast, Youtube}$ represents it performs main-training on YouTube-VOS instead of DAVIS and YouTube-VOS like others}}
\begin{center}
\begin{tabular}{l|ccc|ccc|ccc}
\toprule[1.5pt]
\multirow{2}{*}{Method} & \multicolumn{3}{c|}{VOST val} & \multicolumn{3}{c|}{VOST test} & \multicolumn{3}{c}{\textcolor{black}{VOST val 10fps}} \\ \cline{2-10}
                  & $\mathcal{J}$     & $\mathcal{J}_{tr}$     & $\Delta_{tr}$     & $\mathcal{J}$     & $\mathcal{J}_{tr}$     & $\Delta_{tr}$ & $\mathcal{J}$     & $\mathcal{J}_{tr}$     & $\Delta_{tr}$     \\
\midrule[1pt]
SwinB-AOT-L \cite{AOT_NEURIPS2021_147702db}       & 47.3     & 33.8         & -13.5         & 43.5     & 30.3         & -13.2 & 47.8     & 33.8         & -14.0         \\
SwinB-AOT-L $^{\ast}$ \cite{AOT_NEURIPS2021_147702db}       & 48.4     & 32.9         & -15.5         & 49.7     & 33.9         & -15.8 & 48.3     & 32.5        & -15.8         \\
SwinB-DeAOT-L \cite{DeAOT_yang2022decoupling}     & 46.2     & 32.3         & -13.9         & 49.6     & 35.1         & -14.5 & 47.2     & 34.0         & -13.2         \\
SwinB-DeAOT-L $^{\ast}$ \cite{DeAOT_yang2022decoupling}     & \textcolor{blue}{50.0}     & \textcolor{red}{38.6}         & -11.4         & \textcolor{blue}{51.0}     & \textcolor{blue}{35.7}         & -15.3 & 49.6     & 38.5         & \textcolor{red}{-11.1}         \\
XMem   \cite{XMem_10.1007/978-3-031-19815-1_37}              & 43.6     & 29.2         & -14.4         & 44.5     & 32.8         & \textcolor{red}{-11.7} & 46.1     & 32.5         & -13.6         \\
XMem $^{\ast, Youtube}$  \cite{XMem_10.1007/978-3-031-19815-1_37}              & 44.1     & 33.8         & -10.3         & 44.0     & 32.0     & -12.0 & -     & -         & -         \\
XMem  $^{\ast}$  \cite{XMem_10.1007/978-3-031-19815-1_37}              & 45.9     & 34.2         & -11.7         & 46.7     & 33.2         & -13.5 & 48.6     & 33.9         & -14.7         \\
XMem  $^{\ast \dag}$  \cite{XMem_10.1007/978-3-031-19815-1_37}              & 45.9     & 33.2         & -12.7         & 46.8     & 33.1         & -13.7 & 46.3     & 32.4         & -13.9         \\
\textcolor{black}{AOT $^{\ast, Youtube}$}  \cite{VOST_DBLP:conf/cvpr/TokmakovLG23}              & 48.7 & 36.4 & -12.4         & 49.9     & 37.1         & -12.8 & -     & -         & -         \\
\textcolor{black}{AOT+R-STM  $^{\ast, Youtube}$}  \cite{VOST_DBLP:conf/cvpr/TokmakovLG23}              & 49.7 & 38.5 & \textcolor{red}{-11.2}         & -     & -         & - & \textcolor{blue}{51.9}     & \textcolor{blue}{40.7}         & -11.2         \\
\midrule[1pt]
Ours              & 51.3     & 35.2         & -16.1         & 52.5     & 36.2         & -16.3  & 51.5     & 35.4         & -16.1         \\
Ours  $^{\ast}$              & \textcolor{red}{52.8}     & 36.0         & -16.8         & \textcolor{red}{55.6}     & \textcolor{red}{42.5}         & -13.1 & \textcolor{red}{52.9}     & 36.2         & -16.7         \\
\bottomrule[1.5pt]
\end{tabular}
\end{center}
\label{table:sota_VOST}
\end{table}

\begin{table}[]
\small
\fontsize{9pt}{4mm}\selectfont
\setlength{\tabcolsep}{0.35mm}
\caption{Quantitative evaluation on LVOS validation and test split. \\
The methods in \textcolor{gray}{gray} and black indicate the models trained on DAVIS and Youtube-VOS datasets \textcolor{black}{only} (zero-shot performance), and fine-tuned on LVOS train split\textcolor{black}{, respectively}.
Due to the long length of this dataset, we adjust the memory interval of AOT-L and DeAOT-L to 100 \textcolor{black}{for preventing} GPU memory explosion.
}
\begin{center}
\begin{tabular}{l|cccc|cccc}
\toprule[1.5pt]
\multirow{2}{*}{Method} & \multicolumn{4}{c|}{LVOS val} & \multicolumn{4}{c}{LVOS test} \\ \cline{2-9} 
 & $\mathcal{J}\&\mathcal{F}$  & $\mathcal{J}$     & $\mathcal{F}$     & $\mathcal{V}\downarrow$     & $\mathcal{J}\&\mathcal{F}$  & $\mathcal{J}$     & $\mathcal{F}$    & $\mathcal{V}\downarrow$     \\
\midrule[1pt]
\textcolor{gray}{SwinB-AOT-L  $^{\ast}_{100}$} \cite{AOT_NEURIPS2021_147702db}         & \textcolor{blue}{61.7}     & \textcolor{blue}{56.9}     & \textcolor{blue}{66.5}     & 29.0     & 57.2 & 52.7 & 61.7 & 29.1 \\
\textcolor{gray}{SwinB-DeAOT-L $^{\ast}_{100}$} \cite{DeAOT_yang2022decoupling}        & 60.1     & 55.0     & 65.2     & 29.8     & \textcolor{blue}{57.6} & \textcolor{blue}{53.2} & \textcolor{blue}{61.9} & 29.4 \\
\textcolor{gray}{XMem $^{\ast}$}  \cite{XMem_10.1007/978-3-031-19815-1_37}                   & 46.4 & 42.3 & 50.5 & 28.6 & 49.4 & 44.9 & 54.0  & 29.9 \\
\textcolor{gray}{XMem $^{\ast \dag}$}  \cite{XMem_10.1007/978-3-031-19815-1_37}                 & 54.1 & 50.1 & 58.1 & 29.0  & 49.7 & 45.7 & 53.6 & 31.5 \\
\textcolor{gray}{Cutie $^{\ast}$}  \cite{Cutie}                   & \textcolor{black}{60.1}	& \textcolor{black}{55.9}	& \textcolor{black}{64.2}	& \textcolor{black}{-}	& \textcolor{black}{56.2}	& \textcolor{black}{51.8}	& \textcolor{black}{60.5}	& \textcolor{black}{-} \\
SwinB-AOT-L $^{\ast}_{100}$ \cite{AOT_NEURIPS2021_147702db}         & 55.6 & 51.1 & 60.2 & 28.3 & 58.3 & 53.0 & 63.6 & 27.3  \\
SwinB-DeAOT-L $^{\ast}_{100}$ \cite{DeAOT_yang2022decoupling}        & 56.3 & 51.9 & 60.6 & 31.4 & 54.4 & 49.9 & 59.0 & 28.4 \\
XMem $^{\ast}$  \cite{XMem_10.1007/978-3-031-19815-1_37}                   & 50.7	& 46.3	& 55.2	& 28.3	& 51.1	& 46.7	& 55.6	& \textcolor{red}{25.9} \\
XMem $^{\ast \dag}$  \cite{XMem_10.1007/978-3-031-19815-1_37}                 & 59.0 & 54.5 &	63.4 & \textcolor{red}{27.0} & 51.8 & 47.6 &	56.0 & 28.3 \\
\midrule[1pt]
\textcolor{gray}{Ours $^{\ast}$}        & 63.1     & 58.7     & 67.5     & 31.4     & 59.9     & 55.8     & 64.1     & 27.2     \\
\textcolor{gray}{Ours $^{\ast \dag}$}   & 66.6 & 62.1 & 71.1 & 27.9 & 60.4 & 56.5 & 64.2 & 27.5    \\
Ours $^{\ast}$        & 63.7 & 59.0 & 68.4 & 29.4 & 60.5 & 56.3 & 64.6 & 27.8     \\
Ours $^{\ast \dag}$   & \textcolor{red}{67.1} & \textcolor{red}{62.9} & \textcolor{red}{71.3} & 29.9 & \textcolor{red}{61.4} & \textcolor{red}{57.6} & \textcolor{red}{65.2} & 28.2    \\
\bottomrule[1.5pt]
\end{tabular}
\end{center}
\label{table:sota_LVOS}
\end{table}

\begin{figure*}[pt]
\centering
\includegraphics[width=1.0\linewidth]{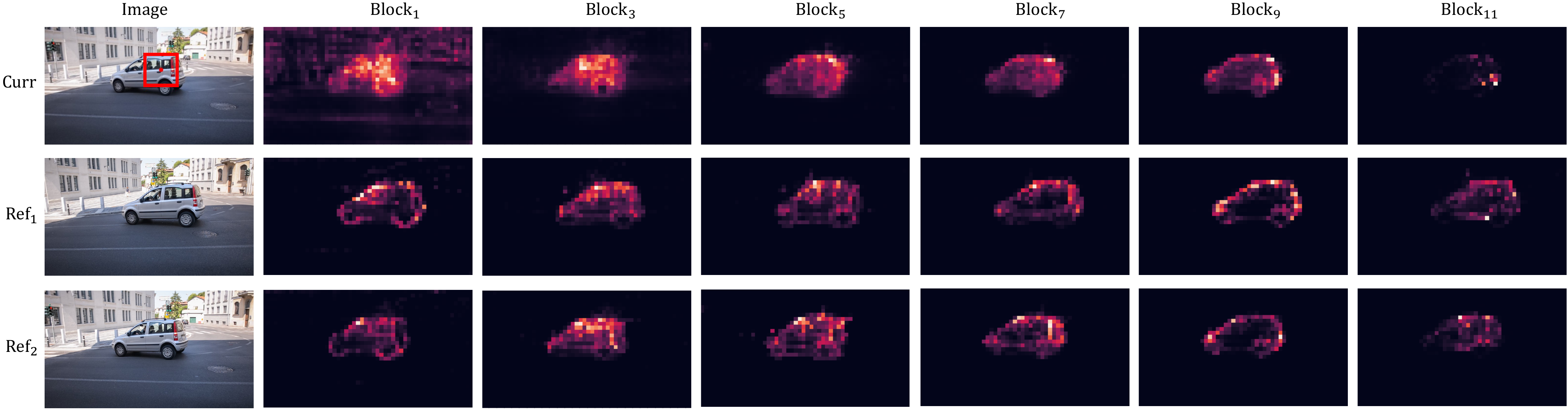}
\caption{
Visualization of attention weights of the current frame in mode (d), corresponding to itself and all reference frames.
We choose the token located in the car (marked with a \textcolor{red}{red} box) of the current frame \textcolor{black}{as the query of attention}.}
\label{fig:attn_currents_self_cross}
\end{figure*}

\noindent \textbf{VOST.} VOST~\cite{VOST_DBLP:conf/cvpr/TokmakovLG23} is a new dataset for \textcolor{black}{transformations}.
Unlike the overall appearance of objects varies little in existing VOS \textcolor{black}{benchmarks, VOST} focuses on segmenting objects as they undergo complex transformations. These transformations preserve virtually nothing of the original except for the identity itself so that it requires more advanced spatio-temporal modeling \textcolor{black}{and holistic understanding as a reliable cue} instead of appearance \textcolor{black}{as usual}.
Due to the transformations, it uses an unambiguous principle: if a region is marked as an object in the ﬁrst frame of a video, all the parts that originate from it maintain the same identity.
\textcolor{black}{It} reports region similarity of the last 25\% of the frames in a sequence additionally to show whether the model can be robust to transformations.
VOST is split into 572 train, 70 validation, and 71 test videos and densely labeled with instance masks.
As shown in Table~\ref{table:sota_VOST}, our model outperforms three SOTA models both on the validation (\textbf{52.9\%}) and test (\textbf{55.6\%}) split, showing that it can accurately segment targets undergoing complex \textcolor{black}{transformations}.
Meanwhile, our model achieves competitive (\textbf{36.2\%} on validation split) and superior (\textbf{42.5\%} on test split) performance in the last 25\% of the frames in a video sequence after the transformation has been mostly completed, and the performance loss $\Delta_{tr}$ is not severe, indicating that our model is robust to \textcolor{black}{the transformations}.

\noindent \textbf{LVOS.} LVOS~\cite{LVOS_hong2022lvos} is a new dataset for  \textcolor{black}{long-term videos}.
According to its author, LVOS is the first densely annotated long-term VOS dataset with high-quality annotations and it is 20 times longer than videos in existing VOS datasets.
Due to its specific property, the difficulty lies in frequent reappearance and cross-temporal similar object confusion.
LVOS has 50 videos both in valid split and test split.
Table~\ref{table:sota_LVOS} shows our model outperforms \textcolor{black}{other} works both on validation and test split with both zero-shot and fine-tuning settings and outperforms their fine-tuning setting with zero-shot setting, which indicates that our model can handle the difficulties of long videos such as long-term reappearing, \textcolor{black}{appearance changing}, and similar objects, even with the simple use of a queue to store reference frames.

\begin{table}[]
\caption{Ablation on propagation modes in Joint Modeling Block. \\
`Refs' and `Current' denote whether the reference tokens accept information from other references and the current tokens, respectively.
}
\begin{center}
\fontsize{9pt}{4mm}\selectfont
\setlength{\tabcolsep}{0.8mm}
\resizebox{\linewidth}{!}{
\begin{tabular}{c|cc|ccc|ccccc}
\toprule[1.5pt]
\multirow{2}{*}{Mode} & \multirow{2}{*}{Refs} & \multirow{2}{*}{Current} & \multicolumn{3}{c|}{DAVIS 2017 test} & \multicolumn{5}{c}{YouTube-VOS 2019 val} \\ \cline{4-11} 
\rule{0pt}{9pt} &                       &                        & $\mathcal{J}\&\mathcal{F}$       & $\mathcal{J}$          & $\mathcal{F}$          & $\mathcal{G}$     & $\mathcal{J}_S$    & $\mathcal{F}_S$    & $\mathcal{J}_U$    & $\mathcal{F}_U$   \\
\midrule[1pt]
na\"{\i}ve                & —  & —          & 76.2       & 72.8       & 79.7       & 82.7  & 83.6  & 88.0 & 75.8   & 83.2 \\
(a)                       & \Checkmark                   & \Checkmark                      & 83.6       & 80.1       & 87.0       & 84.7  & 85.0  & 89.2 & 78.6   & 86.1 \\
(b)                       & \Checkmark                   &                        & 83.8       & 80.3       & 87.3       & 85.2  & 84.1  & 88.5 & \textbf{80.3}   & \textbf{87.7} \\
(c)                       &                   & \Checkmark                      & 83.9       & 80.3       & 87.5       & 85.0  & 84.9  & 89.3 & 79.1  & 86.6 \\ 
\textbf{(d)}                       &                   &                        & \textbf{86.4}       & \textbf{82.7}       & \textbf{90.0}       & \textbf{85.7}  & \textbf{85.6}  & \textbf{90.2}  & 79.8 & 87.2 \\
\bottomrule[1.5pt]
\end{tabular}
}
\end{center}
\label{table:Joint_Blocks_modes}
\end{table}

\begin{table}[]
\caption{Ablation on locations of the Joint Modeling Blocks.}
\centering
\small
\fontsize{9pt}{4mm}\selectfont
\setlength{\tabcolsep}{1.0mm}
\begin{tabular}{c|ccc|ccccc}
\toprule[1.5pt]
\multirow{2}{*}{\begin{tabular}[c]{@{}l@{}}Location\end{tabular}} & \multicolumn{3}{c|}{DAVIS 2017 val} & \multicolumn{5}{c}{Youtube-VOS 2019 val} \\ \cline{2-9} 
\rule{0pt}{9pt} & $\mathcal{J}\&\mathcal{F}$       & $\mathcal{J}$          & $\mathcal{F}$          & $\mathcal{G}$     & $\mathcal{J}_S$    & $\mathcal{F}_S$    & $\mathcal{J}_U$    & $\mathcal{F}_U$  \\
\midrule[1pt]
first 6 blocks  & 87.3       & 84.4       & 90.1      & 84.9  & 84.9  & 89.3  & 78.9 & 86.6 \\
last 6 blocks  & 87.1       & 84.4       & 89.9      & 84.6  & 85.1  & 89.6  & 78.1 & 85.5 \\
evenly 2 blocks  & 86.9       & 83.8       & 89.9      & 85.0  & 85.0  & 89.6  & 78.8 & 86.8 \\
\textbf{all blocks}  & \textbf{88.1}       & \textbf{85.3}       & \textbf{90.9}      & \textbf{85.7}  & \textbf{85.6}  & \textbf{90.2} & \textbf{79.8} & \textbf{87.2} \\
\bottomrule[1.5pt]
\end{tabular}
\label{table:Joint_Blocks_locations}
\end{table}

\noindent \textbf{Qualitative Results.}
We compare our method qualitatively with XMem~\cite{XMem_10.1007/978-3-031-19815-1_37} and Swin-DeAOT~\cite{DeAOT_yang2022decoupling} in Fig.~\ref{fig:results_fig_MOSE_VOST}, all three models are pre-trained with image datasets.
It can be found that our model also performs excellently on the new benchmark, illustrating the generality of our model.
The first and fourth columns show that our model outperforms other models in discriminating long-term, similar objects, while the second and third columns show that our model is better able to handle egocentric videos and drastic transformations.

\subsection{Exploration Studies}
We perform \textcolor{black}{ablation} experiments on DAVIS 2017~\cite{DAVIS17_Pont-Tuset_arXiv_2017} and YouTube-VOS 2019~\cite{youtubeVOS_xu2018youtube} datasets.
By default, we apply ConvMAE with its self-supervised MAE~\cite{MAE_He_2022_CVPR} pre-training, choose propagation mode (d), without using the compressed memory, train only on DAVIS and YouTube-VOS datasets \textcolor{black}{without pre-training on image datasets}, and apply the top-K filter in all blocks for simplicity and fairness.
All the visualizations and qualitative results come from models pre-trained on static image datasets.

\subsubsection{Study on Joint Modeling Blocks}

\noindent \textbf{Study on various propagation modes.} \label{ablation:attention}
To verify the strength of joint modeling inside the backbone, we first implement a na\"{\i}ve baseline that follows the extract-then-matching pipeline and compare it to the joint modeling design with our Joint Modeling Block in four various propagation modes.
Specifically, the baseline follows STCN~\cite{STCN_NEURIPS2021_61b4a64b} to extract \textcolor{black}{key-value pairs for correlation operation} with ConvMAE and ResNet18, calculate affinity matrix and weight sum with dot product instead of L2 similarity to keep consistent with the standard attention of Transformer, and finally pass the aggregated feature into decoder for mask prediction.

As shown in Table~\ref{table:Joint_Blocks_modes}, `Refs' denotes whether the reference tokens share information with other references \textcolor{black}{beyond themselves}, and `Curr' denotes whether they accept information from the current tokens, respectively.
\romannumeral1) Comparing the na\"{\i}ve baseline with our Joint Modeling Block, the results show that joint modeling features and matching in all propagation modes are better than post-matching, which indicates that the decoupled pipeline limits to capture target-specific features at the lower level, \textcolor{black}{which is essential for tracking}.
\romannumeral2) Comparing (a) to (b) or (c) to (d), the reference tokens only perform feature extraction is better than obtaining information from the current tokens.
It demonstrates that since the reference tokens are generated from frames and masks, they can filter out most of the non-target information through the masks \textcolor{black}{while the current tokens contain lots of non-target areas}. Thus, if \textcolor{black}{the current tokens propagate information to reference tokens},
the feature extraction of the references is corrupted by the non-target information \textcolor{black}{from the current tokens severely} before getting sufficient \textcolor{black}{target-specific} features.
\romannumeral3) Comparing (a) to (c) or (b) to (d), each reference frame performs feature extraction separately without sharing with other  \textcolor{black}{references} is better.
It's probably because targets in different reference frames may vary significantly, thus uniformly performing feature extraction destroys the low-level feature in the early stages.
The results also demonstrate our discussion of the difference between SOT and VOS tasks in Section~\ref{sec:relation_with_sot}, \ie, the bidirectional structure is unsuitable for VOS tasks.

\begin{figure}[]
\begin{center}
\includegraphics[width=1.0\columnwidth,keepaspectratio]{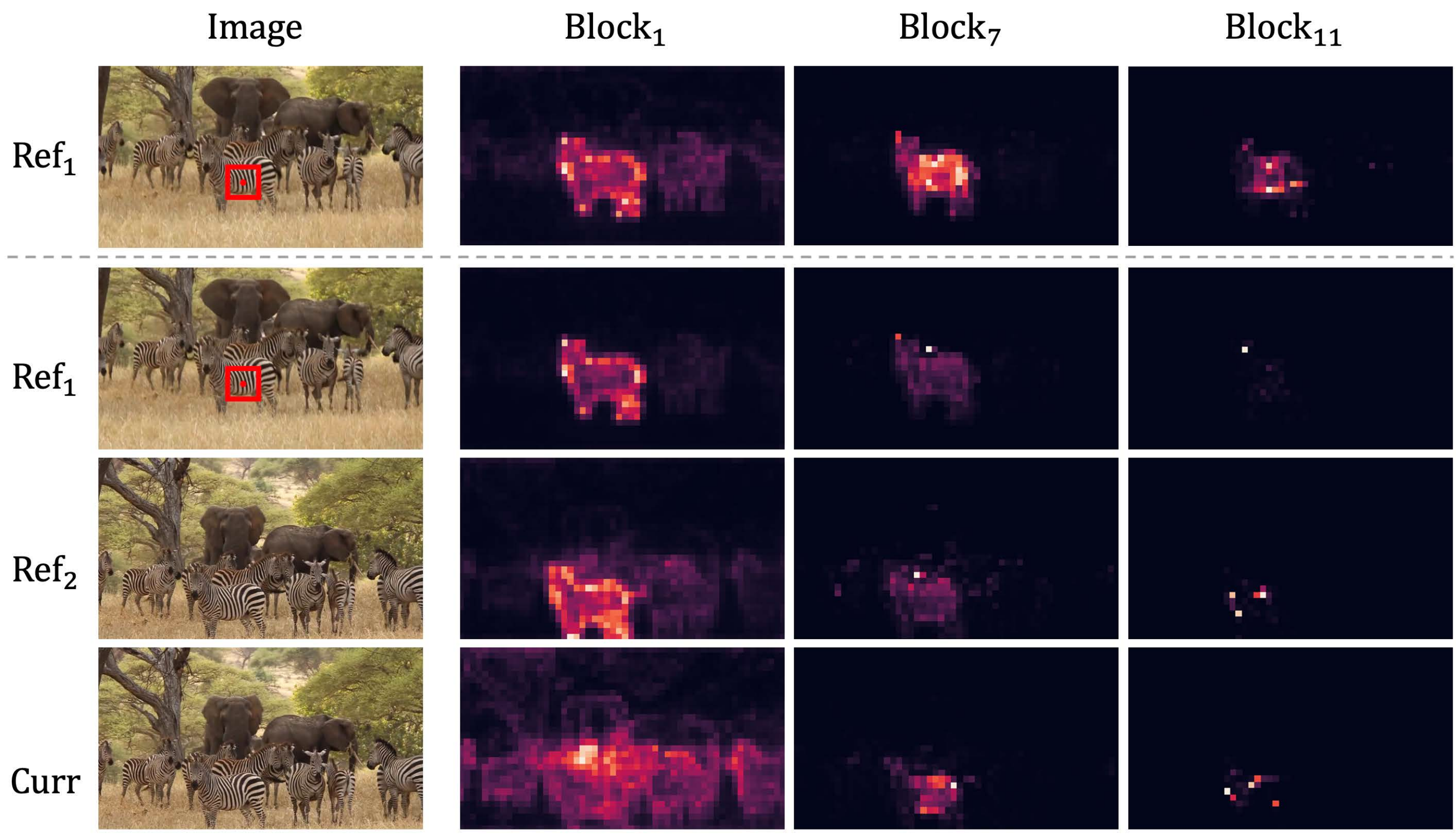}
\end{center}
\caption{
Visualization of attention weights for \textcolor{black}{the \textcolor{red}{red} box area of} one reference frame. The first row is the mode (d) \textcolor{black}{that shows its own impact}.
The last three rows are mode (a), corresponding to itself, another reference, and the current \textcolor{black}{frames}.
}
\label{fig:attn_references_self_cross}
\end{figure}

\begin{figure}[t]
\centering
\includegraphics[width=1.0\columnwidth,keepaspectratio]{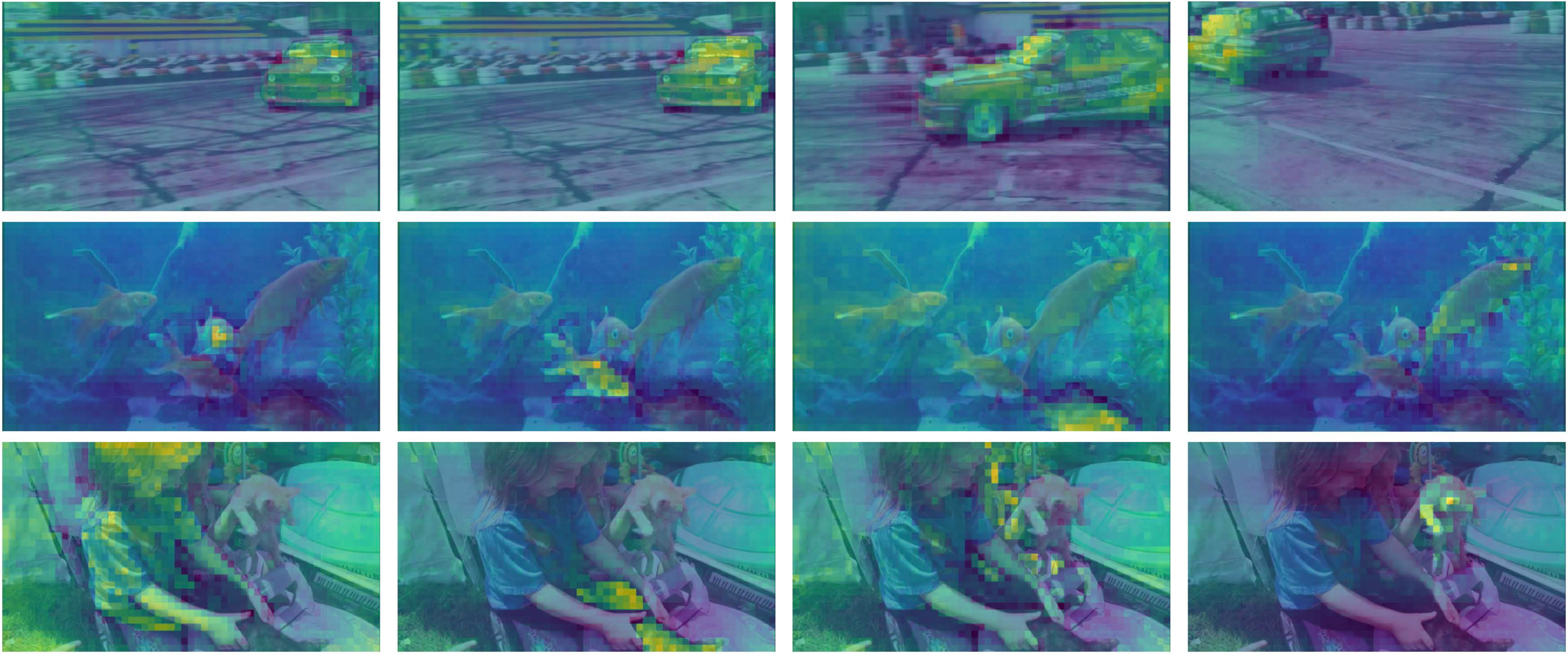}
\caption{
\textcolor{black}{
Visualization of the impact of compressed memory on the current frame.
The first row is the cross-frame effect of a compressed memory on a single target, showing that it can focus well on that fast-moving vehicle.
The second and third rows are the effects of different compressed memory belonging to various targets in the same frame showing its strong discriminatory power.
}
}
\label{fig:attn_clstoken_cross}
\end{figure}

\textcolor{black}{In order to} get a more intuitive sense of the impact of the propagation, we visualize the attention weights \textcolor{black}{of propagating information to the red box area}.
We first visualize the attention weights of the current tokens corresponding to themselves and all reference tokens for mode (d) in Fig.~\ref{fig:attn_currents_self_cross}.
The results show that our model can accurately match the target location of the reference frames.
\textcolor{black}{It pays} more attention to the boundaries in the early stages, and then gradually \textcolor{black}{focuses} on details and critical points \textcolor{black}{in the latter stages}.

In Fig.~\ref{fig:attn_references_self_cross}, we visualize the attention weights \textcolor{black}{of propagating information to the red box area in one reference frame}.
The first row is the feature extraction in mode (d) and the last three rows are \textcolor{black}{in mode (a)}, corresponding to itself, another reference frame, and the current frame.
\textcolor{black}{The visualization shows that in mode (a)} the reference tokens focus on the non-target region of the current a lot, which corrupts the feature extraction of the target \textcolor{black}{so that there is no good achievement of target in reference features}.

\begin{table}[]
\caption{
Ablation on our compressed memory, including \textcolor{black}{comparison with GRU from XMem~\cite{XMem_10.1007/978-3-031-19815-1_37} and the number of blocks for updating compressed memory with the decoder tokens.}
}
\centering
\small
\fontsize{9pt}{4mm}\selectfont
\setlength{\tabcolsep}{1.0mm}
\begin{tabular}{cl|ccc|ccccc}
\toprule[1.5pt]
\multicolumn{2}{c|}{\multirow{2}{*}{Setting}} & \multicolumn{3}{c|}{DAVIS 2017 val}                                                              & \multicolumn{5}{c}{YouTube-VOS 2019 val}                                                                                                                                               \\ \cline{3-10} 
\multicolumn{2}{c|}{}                         & $\mathcal{J}\&\mathcal{F}$     & $\mathcal{J}$                  & $\mathcal{F}$                  & $\mathcal{G}$                  & $\mathcal{J}_S$                & $\mathcal{F}_S$                & $\mathcal{J}_U$                                    & $\mathcal{F}_U$                \\
\midrule[1pt]
\multicolumn{2}{c|}{w/o}                      & 88.1                           & 85.3                           & 90.9                           & 85.7                           & 85.6                           & 90.2                           & 79.8                                               & 87.2                           \\
\multicolumn{2}{c|}{+ GRU \cite{XMem_10.1007/978-3-031-19815-1_37} }                    & 88.1                           & 85.1                           & 91.1                           & 85.9                           & 85.7                           & 90.4                           & 80.0                                               & 87.4                           \\ \hline
\multicolumn{1}{c|}{\multirow{3}{*}{\textbf{+ compressed}}} & 1 & 88.2 & 85.2 & 91.2         & \textbf{86.3} & \textbf{85.8} & 90.4 & 80.4 & \textbf{88.5}  \\
\multicolumn{1}{c|}{} & \textbf{2} & \textbf{89.1} & \textbf{85.9} & \textbf{92.2} & 86.2 & 85.7 & \textbf{90.5} & 80.4 & 88.2 \\
\multicolumn{1}{c|}{} & 3 & 88.1 & 85.1 & 91.0   & 86.2 & 85.7 & 90.5 & \textbf{80.6} & 88.1         \\ 
\bottomrule[1.5pt]
\end{tabular}
\label{table:Compressed_Memory_GRU}
\end{table}

\noindent \textbf{Study on locations of Joint Modeling Block.}\label{ablation:location}
In Table~\ref{table:Joint_Blocks_locations}, we compare the impact from the location of Joint Modeling Blocks.
We set the first six, last six, evenly two, or all blocks as Joint Modeling Blocks, otherwise, it means that the block only performs feature extraction without \textcolor{black}{cross-frame} interaction.
We can find that setting all blocks to Joint Modeling Blocks has the best performance, indicating that both low-level and high-level interactions are necessary.
In addition, setting the first six blocks as Joint Modeling Blocks is better than setting the last six, indicating that interaction at the lower level is important for fine-grained segmentation and discriminative learning, which is missing in previous methods that only \textcolor{black}{perform dense matching at the highest level from backbone output}.

\subsubsection{Study on Compressed Memory}\label{ablation:compressed-memory}

\noindent \textbf{Study on the number of blocks used to interact with decoder tokens.}
In addition to interacting with the reference \textcolor{black}{tokens} from the previous frame, our compressed memory also interacts with the decoder \textcolor{black}{tokens}.
We \textcolor{black}{compared} the performance of various numbers of blocks for interaction with decoder tokens and the results shown in Table~\ref{table:Compressed_Memory_GRU} illustrate the number should be moderate \textcolor{black}{for decent interaction}.
We analyze that fewer blocks may not be sufficient to adequately interact with decoder tokens and thus lose information, while more blocks may instead create an over-reliance on decoder tokens, \textcolor{black}{making it difficult to take full advantage of} reference tokens which are \textcolor{black}{critical components} for joint modeling and online updating. We choose to use 2 blocks \textcolor{black}{to interact} with decoder tokens.

\noindent \textbf{Compare with other memory.}
To verify the \textcolor{black}{robustness} of compressed memory, we compare it to the \textcolor{black}{GRU-based}~\cite{GRU1_cho2014learning, GRU2_cho2014properties} sensory memory from XMem~\cite{XMem_10.1007/978-3-031-19815-1_37}
As shown in Table~\ref{table:Compressed_Memory_GRU}, The experimental results show that it works well both on the short- and long-term datasets, indicating it can provide a more robust target feature and distinguish the target from similar \textcolor{black}{distractors} at the instance level by simply storing one token for each target. This not only illustrates the importance of pre-training from the comparison of GRU which is trained from scratch, but also further states that the modeling of multi-level features inside the backbone and long-term updating are effective.

\begin{table}[]
\caption{Ablation on pretraining methods for backbone.
The `middle', `last', and `PE' represent \textcolor{black}{various structures for building multi-scale features}. \\
See Fig.~\ref{fig:FPN_multi-scale} for details.
}
\centering
\small
\fontsize{9pt}{4mm}\selectfont
\setlength{\tabcolsep}{0.5mm}
\resizebox{\linewidth}{!}{
\begin{tabular}{c|c|ccc|ccccc}
\toprule[1.5pt]
\multirow{2}{*}{Backbone}                                         & \multirow{2}{*}{\begin{tabular}[c]{@{}c@{}}Pre-train\\ Methods\end{tabular}} & \multicolumn{3}{c|}{DAVIS 2017 test} & \multicolumn{5}{c}{YouTube-VOS 2019 val} \\ \cline{3-10} 
\rule{0pt}{9pt} &                                                                                & $\mathcal{J}\&\mathcal{F}$       & $\mathcal{J}$          & $\mathcal{F}$          & $\mathcal{G}$     & $\mathcal{J}_S$    & $\mathcal{F}_S$    & $\mathcal{J}_U$    & $\mathcal{F}_U$   \\ 
\midrule[1pt]
\multirow{4}{*}{\begin{tabular}[c]{@{}c@{}}ViT$_{middle}$ \end{tabular}}  & Scratch                                                                        & 67.8       & 63.9       & 71.8       & 72.7  & 75.0  & 78.4 & 64.8   & 72.7 \\
& DeiT Sup.                                                                      & 77.9       & 74.0       & 81.9       & 81.1  & 81.5  & 85.5 & 75.0   & 82.4 \\
& MAE Ft.                                                                        & 81.0       & 77.1       & 85.0       & 83.0  & 83.3  & 87.3 & 77.1   & 84.5 \\
& MAE Pre.                                                                       & \textbf{81.7}       & \textbf{77.7}       & \textbf{85.8}       & \textbf{84.6}  & \textbf{84.6}  & \textbf{89.1} & \textbf{78.2}   & \textbf{86.5} \\ \hline
\multirow{4}{*}{\begin{tabular}[c]{@{}c@{}}ViT$_{last}$ \end{tabular}} & Scratch                                                                        & 66.7       & 63.0       & 70.5       & 73.1  & 75.6  & 79.0 & 65.2   & 72.7 \\
& DeiT Sup.                                                                      & 78.4       & 74.4       & 82.3       & 82.1  & 81.2  & 85.5 & 76.7   & 85.2 \\
& MAE Ft.                                                                        & 80.5       & 76.5       & 84.6       & 82.7  & 83.0  & 87.5 & 76.4   & 83.9 \\
& MAE Pre.                                                                       & \textbf{82.5}       & \textbf{78.5}       & \textbf{86.5}       & \textbf{85.1}  & \textbf{84.5}  & \textbf{89.0} & \textbf{79.3}   & \textbf{87.4} \\ \hline
\multirow{3}{*}{ConvMAE$_{PE}$}                                 & Scratch & 65.6       & 61.7       & 69.4       & 73.3  & 75.2  & 78.5 & 65.8  & 73.6 \\
& MAE Ft. & 79.7        & 75.8        & 83.7        & 82.5   & 82.8   & 87.1   & 76.0  & 84.1  \\
& MAE Pre. & \textbf{86.4}       & \textbf{82.7}       & \textbf{90.0}       & \textbf{85.7}  & \textbf{85.6}  & \textbf{90.2} & \textbf{79.8}   & \textbf{87.2} \\
\bottomrule[1.5pt]
\end{tabular}
}
\label{table:Backbone_PreTrain}
\end{table}

\begin{table}[]
\caption{Ablation on top-K settings \textcolor{black}{across various backbone networks}. \\
The `Last', `\textcolor{black}{Second} Half', and `All' denote that we use the top-K filter in the last, second half, and all of the blocks, respectively.}
\begin{center}
\footnotesize
\fontsize{9pt}{4mm}\selectfont
\setlength{\tabcolsep}{0.8mm}
\resizebox{\linewidth}{!}{
\begin{tabular}{c|c|ccc|ccccc}
\toprule[1.5pt]
\multirow{2}{*}{Backbone}                                           & \multirow{2}{*}{\begin{tabular}[c]{@{}c@{}}top-K\\ Blocks\end{tabular}} & \multicolumn{3}{c|}{DAVIS 2017 test} & \multicolumn{5}{c}{YouTube-VOS 2019 val} \\ \cline{3-10} 
 \rule{0pt}{9pt}&                             & $\mathcal{J}\&\mathcal{F}$       & $\mathcal{J}$          & $\mathcal{F}$          & $\mathcal{G}$     & $\mathcal{J}_S$    & $\mathcal{F}_S$    & $\mathcal{J}_U$    & $\mathcal{F}_U$   \\ 
\midrule[1pt]
\multirow{3}{*}{\begin{tabular}[c]{@{}c@{}}ViT$_{middle}$ \end{tabular}} & Last                  & \textbf{82.0}       & \textbf{78.1}       & \textbf{86.0}       & 84.3  & 84.6  & 89.1  & 78.0 & 85.6 \\
& \textcolor{black}{Second} Half                  & 81.7       & 77.7       & 85.8       & \textbf{84.6}  & \textbf{84.6}  & \textbf{89.1}  & \textbf{78.2} & \textbf{86.5} \\
& All                   & 81.5       & 77.6       & 85.4       & 84.5  & 84.1  & 88.5  & 78.7 & 86.7 \\ \hline
\multirow{3}{*}{\begin{tabular}[c]{@{}c@{}}ViT$_{last}$ \end{tabular}}   & Last                  & 82.5       & \textbf{78.6}       & 86.4       & 85.0  & 84.3  & 88.9  & \textbf{79.5} & 87.2 \\
& \textcolor{black}{Second} Half                  & \textbf{82.5}       & 78.5       & \textbf{86.5}       & \textbf{85.1}  & \textbf{84.5}  & \textbf{89.0}  & 79.3 & \textbf{87.4} \\
& All                   & 82.4       & 78.4       & 86.4       & 84.5  & 84.1  & 88.7  & 78.7 & 86.6 \\ \hline
\multirow{3}{*}{ConvMAE$_{middle}$}                                            & Last                  & 84.8       & 81.0       & 88.6       & 85.4  & 85.4  & 89.9  & 79.6 & 86.7 \\
& \textcolor{black}{Second} Half                  & 86.1       & 82.2       & 89.9       & \textbf{85.6}  & \textbf{85.6}  & \textbf{90.3}  & \textbf{79.7} & 87.0 \\
& All                   & \textbf{86.3}       & \textbf{82.5}       & \textbf{90.2}       & 85.5  & 85.3  & 90.0  & 79.5 & \textbf{87.2} \\  \hline
\multirow{3}{*}{ConvMAE$_{last}$}                                            & Last                  & 84.8       & 81.0       & 88.5       & 84.7  & 85.2  & 89.8  & 78.4 & 85.5 \\
& \textcolor{black}{Second} Half                  & 85.6       & 81.7       & 89.5       & 85.4  & \textbf{85.2}  & \textbf{90.1}  & 79.2 & 87.0 \\
& All                   & \textbf{86.0}       & \textbf{82.3}       & \textbf{89.8}       & \textbf{85.5}  & 84.9  & 89.7  & \textbf{79.7} & \textbf{87.6} \\  \hline
\multirow{3}{*}{ConvMAE$_{PE}$}                                            & Last                  & 84.4       & 80.7       & 88.0       & 85.3  & 85.6  & 90.1  & 79.3 & 86.4 \\
& \textcolor{black}{Second} Half                  & 86.1       & 82.4       & 89.7       & 85.7  & 85.4  & 90.1  & 79.8 & \textbf{87.5} \\
& All                   & \textbf{86.4}       & \textbf{82.7}       & \textbf{90.0}       & \textbf{85.7}  & \textbf{85.6}  & \textbf{90.2}  & \textbf{79.8} & 87.2 \\
\bottomrule[1.5pt]
\end{tabular}
}
\end{center}
\label{table:Backbone_TopK}
\end{table}

\begin{figure}[t]
\centering
\includegraphics[width=1.0\columnwidth,keepaspectratio]{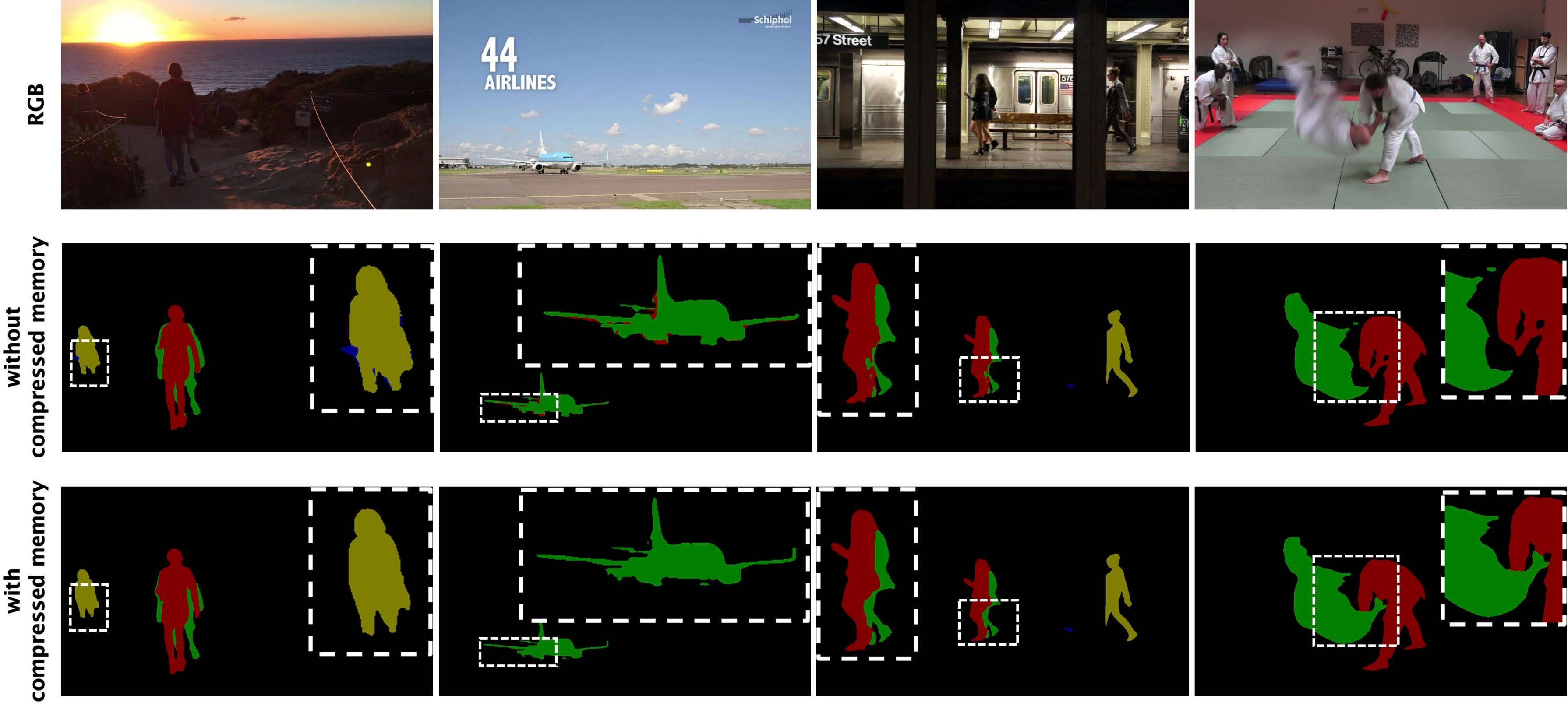}
\caption{
Qualitative comparisons \textcolor{black}{on} compressed memory.
It can be seen that our compressed memory improves the model's discrimination of similar distractors and adapts to \textcolor{black}{the variation of targets}.
}
\label{fig:fig_memory}
\end{figure}

We visualize the attention weights of current tokens corresponding to the compressed memory after \emph{sigmoid} in Fig.~\ref{fig:attn_clstoken_cross} to further explore how the compressed memory works.
It can be noticed that the compressed memory highly responds at the exact position for both deformed long-term single targets and obscured multiple targets, and it exhibits strong discrimination between different targets even without any interaction \textcolor{black}{with other} compressed memory tokens of various targets.
The lower response values for the target boundaries of compressed memory indicate that it can also correct the \textcolor{black}{boundaries of segmentation}. These abilities work well with details and similar objects that are critical for VOS tasks.
Moreover, we provide the results without and with compressed memory in Fig.~\ref{fig:fig_memory} to visualize its effects. We can see that our compressed memory improves the model's ability to discriminate between similar distractors and adapt to variations in the target.

\subsubsection{Study on ViT~\& ConvMAE Architectures}

\noindent \textbf{Study on different backbone and pre-training methods.} 
Since most works leverage supervised pre-training or transformer blocks without pre-training, we further investigate the effect of pre-training methods.
Since ConvMAE does not provide supervised pre-training, we leverage ViT with DeiT-3~\cite{deit3_10.1007/978-3-031-20053-3_30} which supervised \textcolor{black}{classification} pre-training on ImageNet-21K and compare it with self-supervised MAE~\cite{MAE_He_2022_CVPR} pre-training on ImageNet-1K.
We apply the top-K filter in the first half blocks for the ViT backbone.
As shown in Table~\ref{table:Backbone_PreTrain}, pre-training shows necessity on both two backbones.
The self-supervised MAE performs better than supervised DeiT with much less data.
Furthermore, we also test the MAE Ft checkpoint which means the backbone is \textcolor{black}{classification} fine-tuned on ImageNet-1k after MAE pre-trained. The result shows that MAE pre-training is better than fine-tuning, which means \textcolor{black}{the former one} can alleviate the problem that ViT has fewer inductive biases.
We expect the MAE pre-training to contribute to more VOS works \textcolor{black}{as a convenient and effective insight} in the future.

\begin{figure}[t]
\begin{center}
\includegraphics[width=1.0\columnwidth,keepaspectratio]{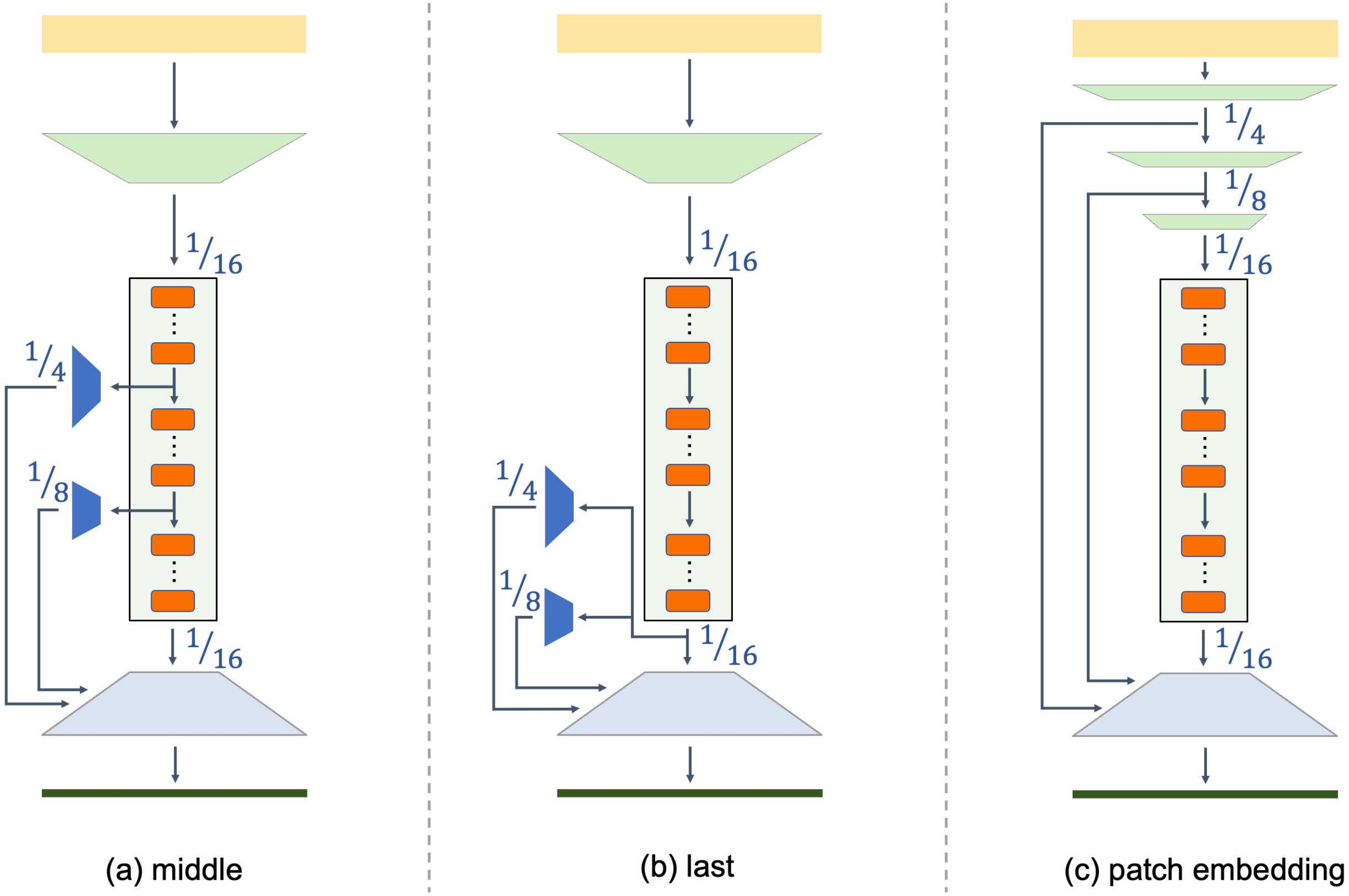}
\end{center}
\caption{\textcolor{black}{
Three structures of building multi-scale features.
The `middle' and `last' represent that we adopt features from the middle blocks or the last blocks of the whole plain backbone with the transposed convolutions for upsampling them.
The `PE' represents that we adopt the features that output from the patch embedding layers.
}
}
\label{fig:FPN_multi-scale}
\end{figure}

\noindent \textbf{Study on the different multi-scale structures.}
We compare various strategies for building multi-scale features \textcolor{black}{which are adopted to predict mask in decoder}.
Specifically, we follow ViTDet~\cite{ViTDet_10.1007/978-3-031-20077-9_17} and implement two constructions to provide the multi-scale features with plain structure by transposed convolutions.
As shown in Fig.~\ref{fig:FPN_multi-scale}, \textcolor{black}{the} `middle' means we divide the backbone into three stages and build the multi-scale features with the outputs from the last block of each stage, and \textcolor{black}{the} `last' means we only use the last block of the whole backbone. For ConvMAE, since its patch embedding layer is multi-stage, we directly use the outputs of patch embedding layers and the last block, denoted as `PE'.
The results show that `patch embedding' is better than `middle' and `last', indicating that the large stride (16) of the patch embedding layer \textcolor{black}{of ViT} makes it \textcolor{black}{lose} spatial details in severe, \textcolor{black}{which is difficult to recover with single-scale features}.
\textcolor{black}{The results of `last' are better than `middle' indicating that} building multi-scale features with the top-level features directly is better than with low-level features.

\begin{table}[]
\caption{Fair comparison on performance and efficiency of various methods with ConvMAE backbone.
The $^{\diamond}$ indicates we replace their backbone with \textcolor{black}{ConvMAE backbone and pre-training}, and \textcolor{black}{train} them on DAVIS and YouTube-VOS datasets. `Ours - T' means we use T reference frames except the previous frame.}
\fontsize{7pt}{3mm}\selectfont
\begin{tabular}{l|cccccccc}
\toprule[1.5pt]
Method & $D_{17}^{val}$ & $D_{17}^{td}$ & $Y_{19}^{val}$ & $Y_{18}^{val}$ & FPS & Mem & \textcolor{black}{Params} \\
\midrule[1pt]
AOT-L$^{\diamond}$   & 80.7 & 74.5 & 78.5 & 78.8 & 4.9 & 17847 & \textcolor{black}{89}            \\
DeAOT-L$^{\diamond}$ & 84.3 & 80.9 & 83.4 & 83.5 & 7.6 & 6144 & \textcolor{black}{94}            \\
XMem$^{\diamond}$  & 86.4 & 84.2 & 85.6 & 85.3 & 25.5 & 1877 & \textcolor{black}{87}            \\
\midrule[1pt]
Ours - 1         & 89.1 & 87.0 & 85.5 & 85.8 & 6.7 & 4404 &\textcolor{black}{108}           \\
Ours - 2         & - & - & - & 86.0 & 3.6 & 5030 & \textcolor{black}{108}           \\
Ours - 3         & - & - & 86.2 & 86.0 & 3.0 & 5785 & \textcolor{black}{108}           \\
\bottomrule[1.5pt]
\end{tabular}

\label{table:backbone-convmae}
\end{table}

\noindent \textbf{Study on top-K filter settings.} \label{ablation:backbone}
In Table~\ref{table:Backbone_TopK}, \textcolor{black}{We compare the impact of various top-K filter settings across different backbone networks.}
The results show that ViT \textcolor{black}{performs its best when} applying the top-K filter in the second half of the blocks while ConvMAE requires it in all blocks.
It suggests that ViT needs more references to add spatial details \textcolor{black}{at low level} due to its aggressive patching embedding layer, but ConvMAE retains \textcolor{black}{detailed feature with} the shallow convolutional layers \textcolor{black}{so that it simply needs to match better in all blocks without worrying about fine-grained features}.

\noindent \textbf{Fair comparison on ConvMAE.}
\textcolor{black}{Due to the inconsistency of the backbone networks among the different models, we replace the backbone networks of the three open-source models~\cite{AOT_NEURIPS2021_147702db,DeAOT_yang2022decoupling,XMem_10.1007/978-3-031-19815-1_37} with the ConvMAE and MAE pre-training that we used.}
We follow their training settings, \textcolor{black}{except for training} them with 160K iterations on DAVIS and YouTube-VOS datasets without synthetic image pre-training, \textcolor{black}{which was the same as ours}.
The comparisons of performance, efficiency, and memory consumption are shown in Table~\ref{table:backbone-convmae}, showing that our model outperforms them substantially with the same backbone, indicating that our performance strengths come from joint modeling and compressed memory design, not just from transformer backbone and \textcolor{black}{MAE} pre-training.

\section{Limitation and Future Work}

Admittedly, \textcolor{black}{the limitation of our model in terms of efficiency is due to the multiple matching in each layer during the joint modeling}.
\textcolor{black}{To verify the impact of reducing the number of matching objects on performance and efficiency, we try to reduce} the number of reference frames. \textcolor{black}{As shown in Table~\ref{table:backbone-convmae}}, the performance degradation is small while significantly improving inference speed, allowing our model to achieve similar speeds as \cite{AOT_NEURIPS2021_147702db} and \cite{DeAOT_yang2022decoupling} with same backbone.
In the future, we consider the following three enhancement ideas:
\romannumeral1) How to improve model efficiency with as little loss of performance as possible, such as reducing the number of tokens for matching \textcolor{black}{with token pruning} technology~\cite{DBLP:conf/nips/RaoZLLZH21, DBLP:journals/corr/abs-2308-04657, DBLP:conf/aaai/XuZZSLDZXS22}, or reducing the number of layers for joint modeling \textcolor{black}{with layer pruning~\cite{MixFormerV2}}.
\romannumeral2) How to communicate between different targets to improve discriminability, rather than processing them in parallel without interaction.
\romannumeral3) How to measure reference frames and masks stored in the memory bank, rather than using a simple queue.

\section{Conclusions}
This paper proposes JointFormer, a unified framework that jointly models feature, correspondence, and a compressed memory inside the backbone.
As the core module of our JointFormer, Joint Modeling Blocks perform iterative joint modeling to propagate the target information and achieve interaction between current and reference frames.
Due to the specificity of the VOS task, we present various modes for propagating target information and demonstrate the asymmetric structure is better than bidirectional \textcolor{black}{structure}.
Furthermore, we develop a customized online update mechanism for our compressed memory to provide a long-term and adaptive representation.
Extensive experiments show that our JointFormer obtains a notable improvement over previous works on the most dominant benchmarks and all four new benchmarks with various difficulties which illustrates the strength and robustness of our JointFormer.
We hope that joint modeling will benefit future research on VOS task, and our JointFormer can serve as a powerful baseline for multiple benchmarks.

\section{Acknowledgements}
This work was supported by the National Key R$\&$D Program of China under Grant 2022ZD0160900, in part by the Fundamental Research Funds for the Central Universities under Grant 020214380119, in part by the Jiangsu Frontier Technology Research and Development Program under Grant BF2024076, and in part by the Collaborative Innovation Center of Novel Software Technology and Industrialization.




\ifCLASSOPTIONcaptionsoff
  \newpage
\fi

{\small

\bibliographystyle{ieee_fullname}
\bibliography{main}
}


\begin{IEEEbiography}[{\includegraphics[width=1in,height=1.25in,clip,keepaspectratio]{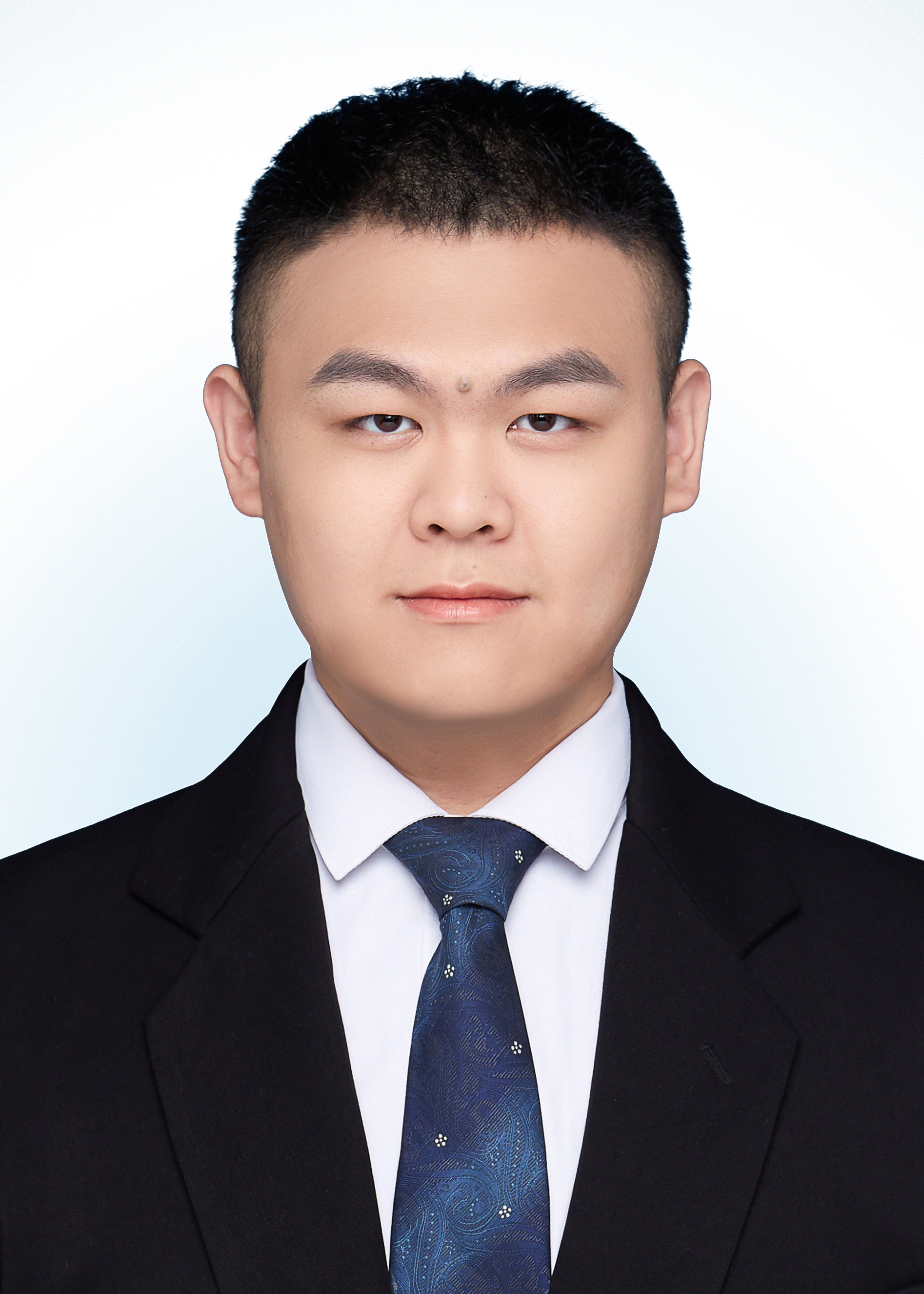}}]
 {Jiaming Zhang} received the BSc degree from Beijing Institute of Technology, Beijing, China, in 2021. He is currently working toward the PhD degree with the Department of Computer Science and Technology, Nanjing University.
 His research interests include computer vision and deep learning, with a focus on video segmentation and video understanding.
\end{IEEEbiography}

\begin{IEEEbiography}[{\includegraphics[width=1in,height=1.25in,clip,keepaspectratio]{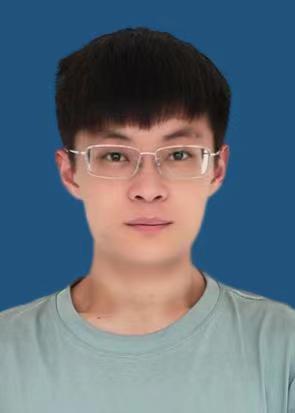}}]
 {Yutao Cui} received the B.Sc. and M.S. degree from Beijing Institute of Technology, Beijing, China, in 2017 and 2019 respectively. He
is currently pursuing the Ph.D. degree of the Department of Computer Science and Technology, Nanjing University, Nanjing, China. His current research interests include visual object tracking and video object segmentation.
\end{IEEEbiography}

\begin{IEEEbiography}[{\includegraphics[width=1in,height=1.25in,clip,keepaspectratio]{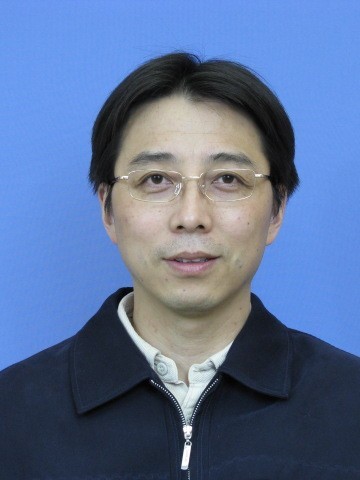}}]
 {Gangshan Wu} (Member, IEEE) received the BSc, MS, and PhD degrees from the Department of Computer Science and Technology, Nanjing University, Nanjing, China, in 1988, 1991, and 2000 respectively. He is currently a professor with the Department of Computer Science and Technology, Nanjing University. His current research interests include computer vision, multimedia content analysis, multimedia information retrieval, digital museum, and large-scale volumetric data processing.
\end{IEEEbiography}

\begin{IEEEbiography}[{\includegraphics[width=1in,height=1.25in,clip,keepaspectratio]{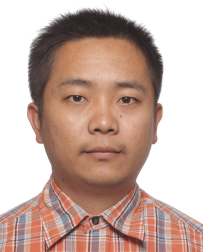}}]
 {Limin Wang} (Senior Member, IEEE) received the B.Sc. degree from Nanjing University, Nanjing, China, in 2011, and the Ph.D. degree from the Chinese University of Hong Kong, Hong Kong, in 2015. From 2015 to 2018, he was a Post-Doctoral Researcher with the Computer Vision Laboratory, ETH Zurich. He is currently a Professor with the Department of Computer Science and Technology, Nanjing University. His research interests include computer vision and deep learning. He has served as an Area Chair for NeurIPS, CVPR, ICCV, and is on the Editorial Board of IJCV and PR.
\end{IEEEbiography}

\end{document}